\documentclass[10pt,twocolumn,letterpaper]{article}

\usepackage{cvpr}
\usepackage{graphicx}
\usepackage{amsmath}
\usepackage{amssymb}
\usepackage{booktabs}
\usepackage{bbm}
\usepackage{algorithm,algpseudocode}
\usepackage[export]{adjustbox}
\algnewcommand{\Inputs}[1]{%
  \State \textbf{Inputs:}
  \Statex \hspace*{\algorithmicindent}\parbox[t]{.8\linewidth}{\raggedright #1}
}
\algnewcommand{\Outputs}[1]{%
  \State \textbf{Outputs:}
  \Statex \hspace*{\algorithmicindent}\parbox[t]{.8\linewidth}{\raggedright #1}
}
\algnewcommand{\Initialize}[1]{%
  \State \textbf{Initialize:}
  \Statex \hspace*{\algorithmicindent}\parbox[t]{.8\linewidth}{\raggedright #1}
}
\usepackage{algpseudocode}
\usepackage{amsmath}
\usepackage[pagebackref,breaklinks,colorlinks]{hyperref}
\usepackage{multirow}
\usepackage{array}
\newcolumntype{P}[1]{>{\centering\arraybackslash}p{#1}}

\usepackage[capitalize]{cleveref}
\crefname{section}{Sec.}{Secs.}
\Crefname{section}{Section}{Sections}
\Crefname{table}{Table}{Tables}
\crefname{table}{Tab.}{Tabs.}

\usepackage{mathptmx}
\usepackage{etoolbox}
\usepackage{amsmath,bm}
\usepackage{xcolor}
\usepackage{amsmath}
\usepackage{amsthm}
\usepackage{mathtools}
\usepackage{bm}
\usepackage{commath}
\usepackage[accsupp]{axessibility}  

\theoremstyle{plain}

\crefname{assumption}{Assump.}{Assumption}

\crefname{definition}{Def.}{Definition}

\crefname{theorem}{Theorem}{Theorems}

\crefname{remark}{Remark}{Remarks}

\crefname{lemma}{Theorem}{Lemmas}

\crefname{proposition}{Theorem}{Propositions}

\crefname{corollary}{Theorem}{Corollaries}

\crefname{example}{Example}{Examples}

\crefname{claim}{Theorem}{Claim}

\def\cvprPaperID{5022} 
\def\confName{CVPR}
\def\confYear{2023}

\begin{document}

\title{BASiS: Batch Aligned Spectral Embedding Space}

\author{Or Streicher $\quad$Ido Cohen $\quad$Guy Gilboa\\
Viterbi Faculty of Electrical and Computer Engineering\\ 
Technion - Israel Institute of Technology, Haifa, Israel\\
{\tt\small orr.shtr@gmail.com  $\quad$ido.coh@gmail.com $\quad$guy.gilboa@ee.technion.ac.il}
}

\maketitle

\begin{abstract}
    Graph is a highly generic and diverse representation, suitable for almost any data processing problem. Spectral graph theory has been shown to provide powerful algorithms, backed by solid linear algebra theory. It thus can be extremely instrumental to design deep network building blocks with spectral graph characteristics. For instance, such a network allows the design of optimal graphs for certain tasks or obtaining a canonical orthogonal low-dimensional embedding of the data. Recent attempts to solve this problem were based on minimizing Rayleigh-quotient type losses. We propose a different approach of directly learning the graph's eigensapce. A severe problem of the direct approach, applied in batch-learning, is the inconsistent mapping of features to eigenspace coordinates in different batches. 
    We analyze the degrees of freedom of learning this task using batches and propose a stable alignment mechanism that can work both with batch changes and with graph-metric changes. We show that our learnt spectral embedding is better in terms of NMI, ACC, Grassman distnace, orthogonality  and classification accuracy, compared to SOTA. In addition, the learning is more stable.

\end{abstract}


\section{Introduction}
\label{sec:intro}
Representing information by using graphs and analyzing their spectral properties has been shown to be an effective classical solution in a wide range of problems including clustering \cite{bresson2013multiclass, ng2001spectral, zelnik2004self}, classification \cite{garcia2014multiclass},  segmentation \cite{shi2000normalized}, dimensionality reduction \cite{belkin2003laplacian, coifman2006diffusion, roweis2000nonlinear} and more. In this setting,  data is represented by nodes of a graph, which are embedded into the eigenspace of the graph-Laplacian, a canonical linear operator measuring local smoothness.

Incorporating  analytic data structures and methods within a deep learning framework has many advantages. It yields better transparency and understanding of the network, allows the use of classical ideas, which were thoroughly investigated and can lead to the design of new architectures, grounded in solid theory.
Spectral graph algorithms, however, are hard to incorporate directly in neural-networks since they require eigenvalue computations which cannot be integrated in back-propagation training algorithms. Another major drawback of spectral graph tools is their low scalability. It is not feasible to hold a large graph containing millions of nodes and to compute its graph-Laplacian eigenvectors. Moreover, updating the graph with additional nodes is combersome and one usually resorts to graph-interpolation techniques, referred to as Out Of Sample Extension (OOSE) methods.

An approach to solve the above problems using deep neural networks (DNNs), firstly suggested in \cite{shaham2018spectralnet} and recently also in \cite{chen2022specnet2}, is to train a network that approximates the eigenspace by minimizing Rayleigh quotient type losses. The core idea is that the Rayleigh quotient of a sum of $n$ vectors is minimized by the $n$ eigenvectors with the corresponding $n$ smallest eigenvalues. As a result, given the features of a data instance (node) as input, these networks generate the respective coordinate in the spectral embedding space. This space should be equivalent in some sense to the analytically calculated graph-Laplacian eigenvector space. A common way to measure the equivalence of these spaces is using the Grassman distance. Unfortunately, applying  this  indirect approach does not guarantee convergence to the desired eigenspace and therefore the captured might not be faithful.

An alternative approach, suggested in \cite{mishne2019diffusion} for computing the diffusion map embedding, is a direct supervised method. The idea is to compute the embedding analytically, use it as ground-truth and train the network to map features to eigenspace coordinates in a supervised manner. In order to compute the ground truth embedding, the authors used the entire training set. This operation is very demanding computationally in terms of both memory and time and is not scalable when the training set is very large.

\begin{figure*}[htbp!]
  \centering
   \includegraphics[trim = 2mm 80mm 2mm 50mm, clip=true,width=1\textwidth]{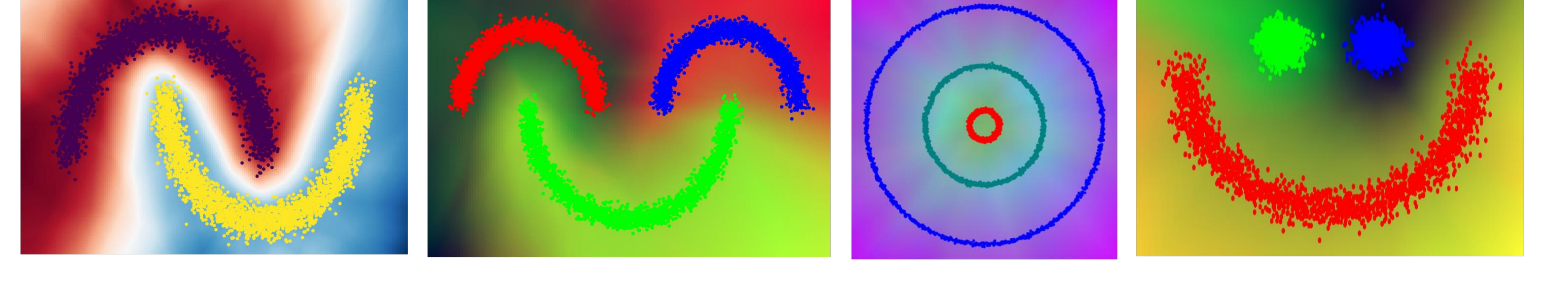}
   \caption{{\bf Toy examples}.  An Illustration of trained BASiS models over common spectral-clustering toy examples. Each figure describes the embedding values given by the network to each point in the space and the clustering results over selected points. BASiS performs successful clustering and is able to interpolate and extrapolate the training data smoothly.}
   \label{fig:model_output_all}
\end{figure*}

Our proposed method is  to learn directly the eigenspace in batches. We treat each batch as sampling of the full graph and learn the eigenvector values in a supervised manner. A major problem of this kind of scheme is the inconsistency in the embedding coordinates. Thus, two instances in different batches with the same features can be mapped to very different eigenspace coordinates. Our solution is to use affine registration techniques to align the batches. Further, we use this alignment strategy to also allow changes in the graph affinity metric.  
Our proposed method retains the following main qualities: 1) {\bf Scalability.} Data is learnt in batches, allowing a training based on large and complex input sets;  2) {\bf OOSE.} Out of sample extension is immediate. 3) {\bf High quality approximation of the eigenspace.} Since our learning method is direct and fully supervised, an excellent approximation of the graph eigenspace is obtained. 4) {\bf Robustness to features change.} We can train the model also when features and affinities between nodes change;
All the above properties yield a spectral building block which can be highly instrumental in various deep learning algorithms, containing an inherent orthogonal low dimensional embedding of the data, based on linear algebraic theory.

\section{Settings and Notations}
Let $\{x_i\}_{i=1}^n$ be a set of data instances denoted as $X$ which is a finite set in $\mathbb{R}^d$. These samples are assumed to lie on a lower dimensional manifold $\mathcal{M}$.

These instances are represented as nodes on an undirected weighted graph $G = (V, E, W)$, where $V$ and $E$ are sets of the vertices and edges, respectively, and $W$ is the adjacency matrix. This matrix is symmetric and defined by a distance measure between the nodes. For example, a common choice is a Gaussian kernel and Euclidean distance,
\begin{equation}\label{Gaussian kernel}
W_{ij}=\exp\left(-\frac{||x_i-x_j||_2^2}{2\sigma^2}\right),
\end{equation}
where $\sigma$ is a soft-threshold parameter.

The degree matrix $D$ is a diagonal matrix where $D_{ii}$ is the degree of the $i$-th vertex, i.e., $D_{ii}=\sum_{j}{W_{ij}}$. 
The graph-Laplacian operator is defined by,
\begin{equation}\label{Laplacian_def}
L := D-W .
\end{equation}
The graph-Laplacian is a symmetric, positive semi-definite matrix, its eigenvalues are real, and its eigenvectors form an orthogonal basis. 
The eigenvalues of $L$ are sorted in ascending order $\lambda_1 \leq \lambda_2 \leq ... \leq \lambda_n$, where the corresponding eigenvectors are denoted by $u_1,u_2...,u_n$. The sample $x_i$ is represented in the spectral embedding space as the $i$th row of the matrix $U=\begin{bmatrix}
    u_1&\cdots&u_K
\end{bmatrix} \in \mathbb{R}^{n \times K}$, denoted as $\varphi_i$. Thus, more formally, the dimensionality reduction process can be formulated as
\begin{equation}\label{embedding_def}
x_i \longmapsto \varphi_i = [u_1(i), u_2(i),..., u_K(i)]\in \mathbb{R}^K,
\end{equation}
where $K\ll d$.
This representation preserves well essential data information
\cite{coifman2006diffusion, katz2019alternating, lederman2018learning, ortega2018graph}

Alternatively, one can replace the Laplacian definition \eqref{Laplacian_def} with
\begin{equation}\label{normalized laplacian}
L_N := D^{-\frac{1}{2}}LD^{-\frac{1}{2}}=I-D^{-\frac{1}{2}}WD^{-\frac{1}{2}}.
\end{equation}
This matrix may yield better performances for certain tasks and datasets
\cite{shi1998motion, shi2000normalized, tatiraju2008image}.


\section{Related Work}
\label{sec:related_work}
OOSE and scalability of graph-based learning methods are ongoing research topics. Mathematical analyses and analytical solutions to these problems can be found, for example, in  \cite{alzate2008multiway, belabbas2009landmark, bengio2003out, fowlkes2004spectral, williams2000using}. However, neural networks learning the latent space of the data usually yield an efficient, robust and reliable  solution for these problems. 
 Moreover, neural network modules can be easily integrated in larger networks, employing this embedding. For a recent use of learnable graphs in semi-supervised learning and data visualization see \cite{aviles2022graphxcovid}.
The effectiveness of modeling PDE's and certain eigenproblems in grid-free, mesh-free manner was  shown in \cite{bar2021strong,weinan2021algorithms,ben2020solving}. 
We review below the main advances in eigenspace embedding. 

 
 

{\bf Diffusion Nets \cite{mishne2019diffusion}.} 
Diffusion Maps (DM) is a spectral embedding, resulting from the eigendecomposition of
\begin{equation}\label{eq:random_walk_matrix}
P := WD^{-1},
\end{equation}
known as the random-walk matrix \cite{coifman2006diffusion}. More formally, similarly to \cref{embedding_def}, diffusion maps is defined by
\begin{equation}\label{dm_embedding_def}
x_i \longmapsto \varphi_i = [\gamma_1^t\Phi_1(i), \gamma_2^t\Phi_2(i),...,\gamma_K^t\Phi_K(i)]\in \mathbb{R}^K,
\end{equation}

where $\{\Phi_j\}_{j=1}^K$ are the first non-trivial eigenvectors of $P$, $\{\gamma\}_{i=j}^K$ are the corresponding eigenvalues and $t>0$ is the diffusion time. Note, that $P$ and $L_N$ have the same eigenvectors, in reverse order with respect to their eigenvalues.

Diffusion Net (DN) is an autoencoder trained to map between the data and the DM. 
The loss function of the encoder is defined by,
\begin{equation}\label{diffusion_net_encoder_loss}
\mathcal{L}_{DN}^e(\theta^e)=\frac{1}{2n}\sum_{i=1}^n\norm{f_{\theta^e}^e(x_i)-\phi_i}^2+ F(\theta^e, X),
\end{equation}
where $\theta^e$ denotes the encoder's parameters,  
$f_{\theta^e}^e(x_i)$ is the encoder output and $F(\theta^e, X) = \frac{\mu}{2}\sum_{l=1}^{L-1}{\norm{\theta^e{_l}}_F^2}+\frac{\eta}{2m}\sum_{j=1}^d{||(P-\gamma_jI)(o^e_j)^T||^2}$ is a regularization term such that $\theta^e{_l}$ are the weights of the $l$-th layer, $o^e_j$ is the $j$-th column of the output matrix, $\mu$ and $\eta$ are regularization parameters. Note, 
Diffusion Net requires 
to compute the embedding of the training set in advance, meaning \emph{it cannot be trained with mini-batches} and therefore has difficulty dealing with large datasets.

{\bf SpectralNet1 \cite{shaham2018spectralnet} (SpecNet1).} 
This DNN learns the embedding corresponds to $L$ by minimizing the \emph{ratio-cut} loss of Ng \etal \cite{ng2001spectral}, without adding an orthogonality constraint on the solution, with the loss
\begin{equation}\label{spectralnet_loss}
\mathcal{L}_{SN1}(\theta) = \frac{1}{m^2}\sum_{i,j=1}^m{W_{i,j}||y_i-y_j||^2} = \frac{2}{m^2}\textrm{tr}(Y^TLY),
\end{equation}
where $y_i=f_{\theta}(x_i)$ is the network output, $m$ is the batch size, and tr is the trace operator. In order to calculate the eigenvectors of $L_N$, one should normalize $y_i, y_j$ with the corresponding node degree.  In SpectralNet1 orthogonality of the training is gained by defining the last layer of the network as a linear layer set to orthogonalize the output. The last layer's weights are calculated during training with QR decomposition over the DNN's outputs. The authors point out that in order to get good generalization and approximate orthogonal output at inference, large batches are required.

{\bf SpectralNet2 \cite{chen2022specnet2} (SpecNet2).} 
In this recent work the authors suggested to  solve the eigenpair problem of the matrix pencil $(W, D)$.
The loss function is defined by,

\begin{equation}\label{specnet2_loss}
\mathcal{L}_{SN2}(\theta) = \frac{1}{m^2}\textrm{tr}\left(-2Y^TW Y+ \frac{1}{m^2}Y^TDYY^TDY\right),
\end{equation}
where $Y$ is the network's output.
Given the output $Y$, an approximation to the eigenvectors of $P$, \cref{eq:random_walk_matrix}, can be calculated as $\hat{U}=YO$ where $O \in \mathbb{R}^{K \times K}$ satisfies
\begin{equation}\label{specnet2_ev_mat_calc}
\begin{aligned}
Y^TWYO = Y^TDYO\Lambda,
\end{aligned}
\end{equation}
where $\Lambda$ is a refined approximation of the eigenvalue  matrix of $(W, D)$. 
Note that \cref{specnet2_ev_mat_calc} requires a batch for its computation, which may be problematic at inference.
The authors 
show qualitatively a successful approximation to the analytical embedding. 

\begin{figure}[htbp!]
\captionsetup[subfigure]{justification=centering}
\centering
    \begin{subfigure}[t]{0.235\textwidth}
        \includegraphics[trim = 10mm 50mm 30mm 60mm, clip,width=1\textwidth,valign=t]{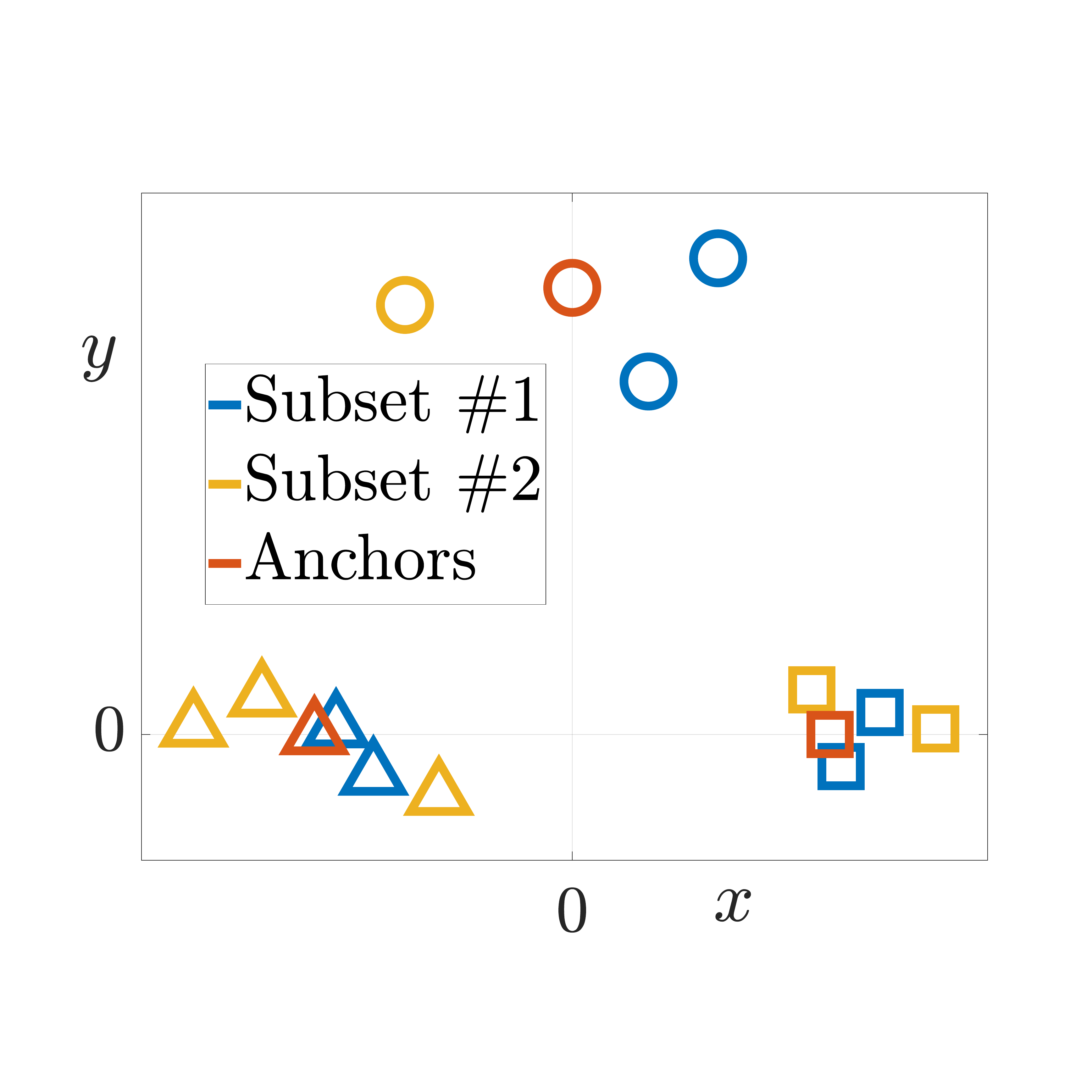}
        \caption{ }
        \label{subfig:toyExmple_data}
    \end{subfigure}
     \begin{subfigure}[t]{0.235\textwidth}
        \includegraphics[trim = 10mm 50mm 30mm 60mm, clip,width=1\textwidth,valign=t]{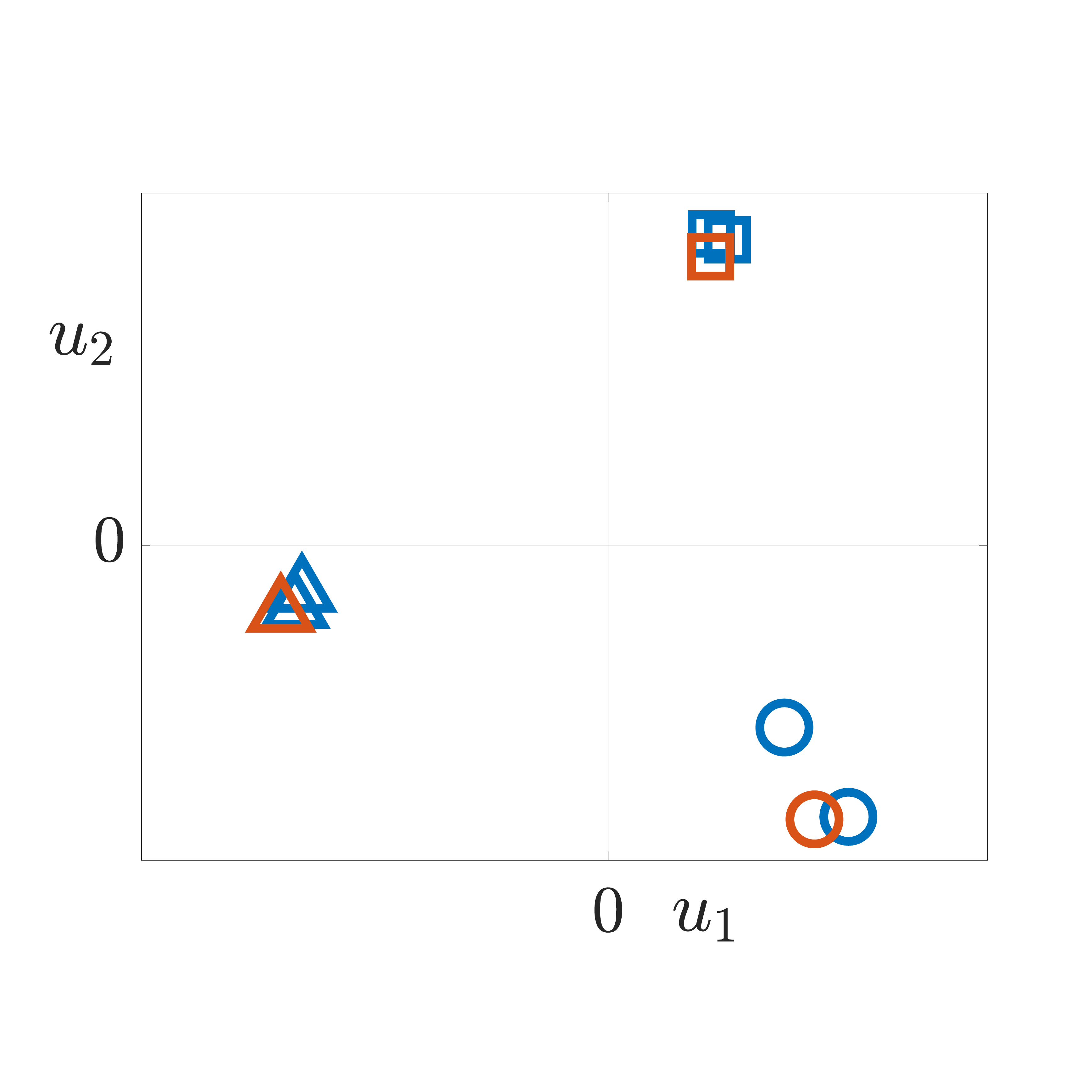}
        \caption{}
        \label{fig:toyExmple_LS1}
    \end{subfigure}\\
     \begin{subfigure}[t]{0.235\textwidth}
        \includegraphics[valign=t, trim = 10mm 50mm 30mm 60mm, clip,width=1\textwidth]{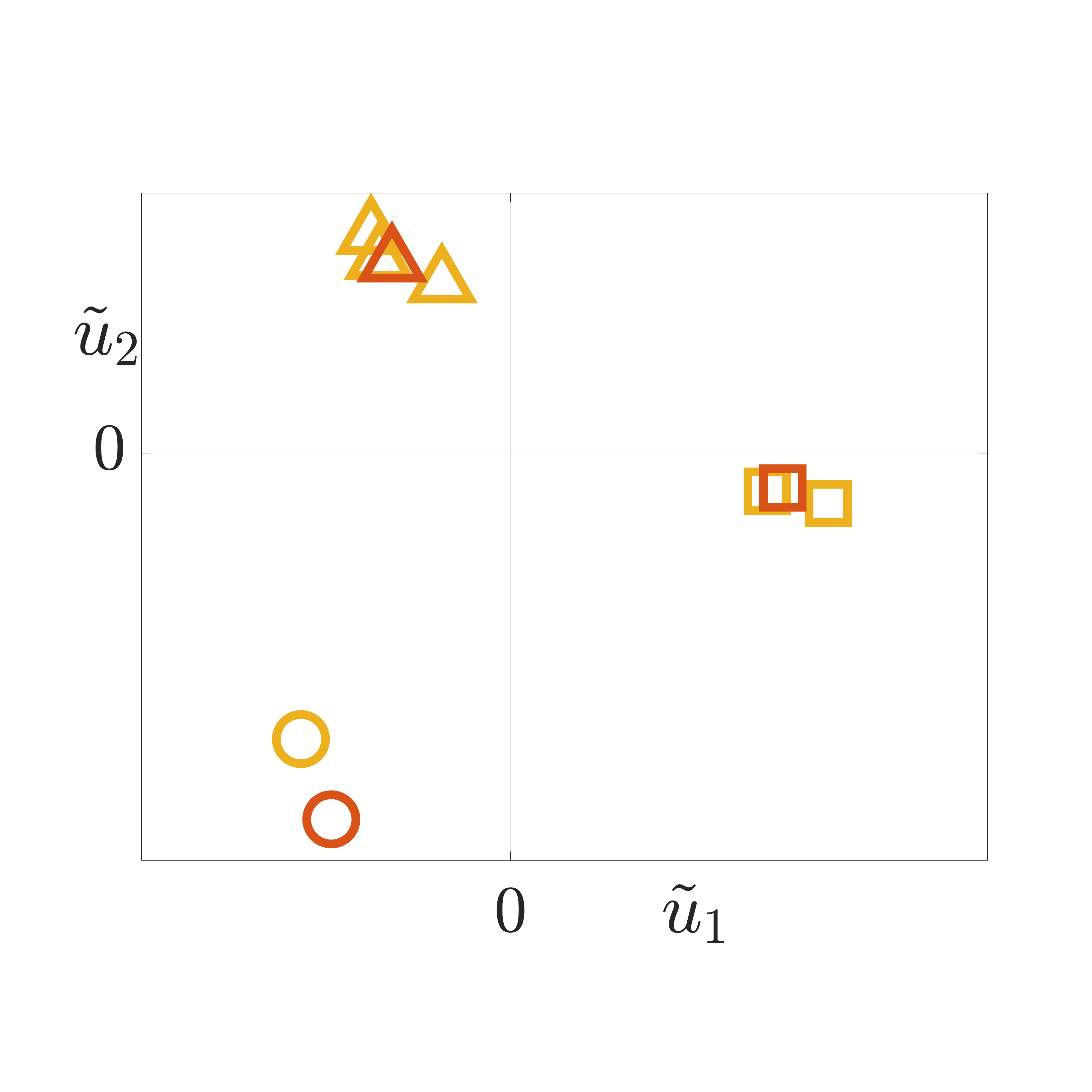}
        \caption{}
        \label{fig:toyExmple_LS2}
    \end{subfigure}
    \begin{subfigure}[t]{0.235\textwidth}
        \includegraphics[valign=t,trim = 10mm 50mm 30mm 60mm, clip,width=1\textwidth]{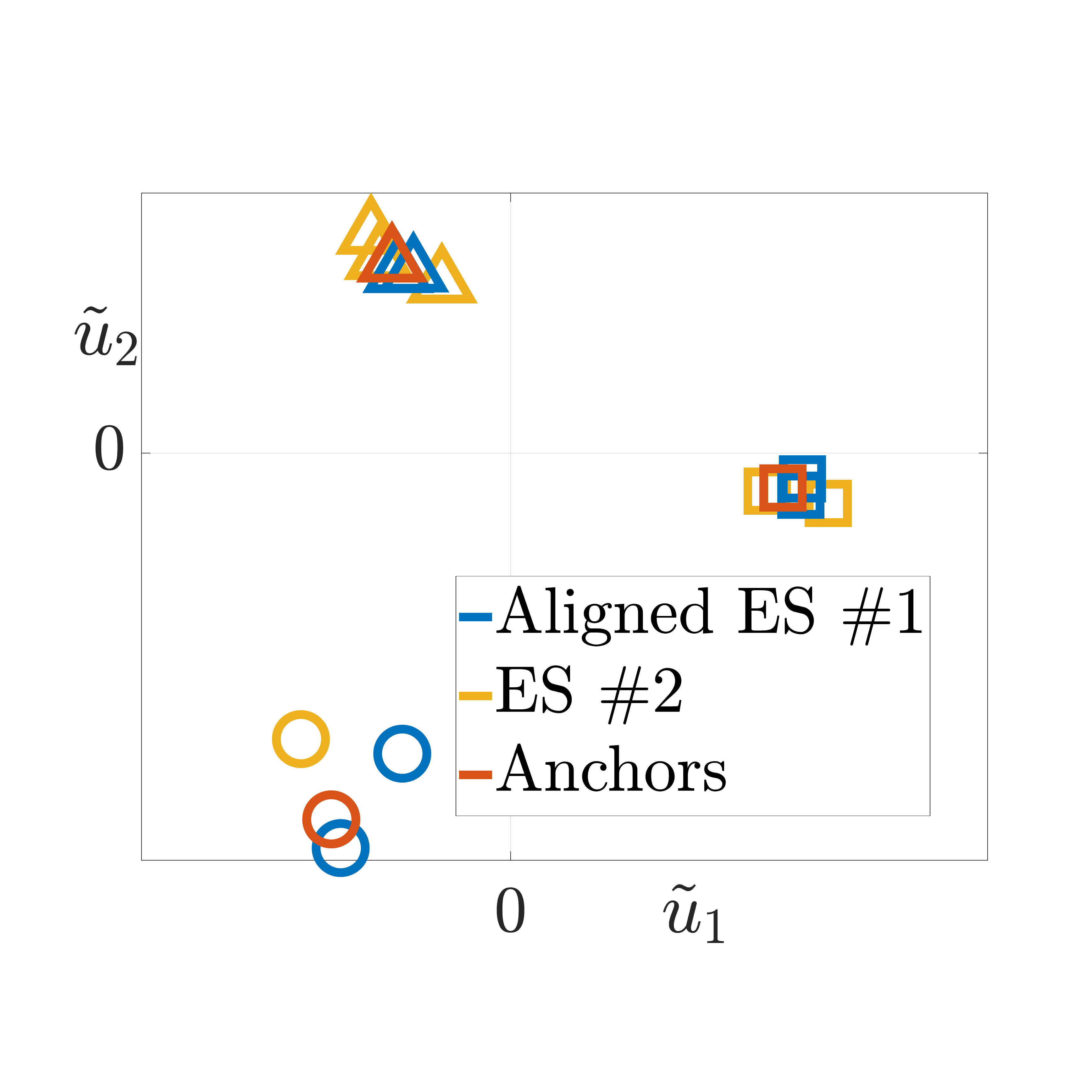}
        \caption{}
        \label{fig:toyExmple_LS2wthLS1}
    \end{subfigure}
    \caption{{\bf Illustration}. The full dataset of three separated clusters, divided into two subsets and anchors is shown in \cref{subfig:toyExmple_data}. Figs. \ref{fig:toyExmple_LS1}-\ref{fig:toyExmple_LS2} show the embedding spaces of subset $\#1$ and subset $\#2$, respectively. \cref{fig:toyExmple_LS2wthLS1} shows the embedding space of the entire data after aligning the embedding of subset $\#1$ to that of subset $\#2$.
    }
    \label{idosfigs:toyExample}
\end{figure}

\section{Our Method}
\subsection{Motivation}
Our goal is to learn a model $f: \mathcal{M} \rightarrow \mathbb{R}^K $, where given a sample $x \in \mathcal{M}$ approximates well the corresponding $\varphi$ of \cref{embedding_def}.
As is common with DNN learning, given a large training set, we would like to train the model by splitting the dataset into batches. A batch can be viewed as sampling the graph of the training set. 
A straightforward approach would be to compute the eigenspace of each batch and to learn a mapping from $x$ to $ \varphi$, using a data loss similar to Diffusion Nets. The problem is that different samples of the training set most often lead to different embeddings. Specifically, the same instance $x_i$ can be mapped very differently in each batch. 

This can be demonstrated in a very simple toy example, shown in \cref{idosfigs:toyExample}, which illustrates the core problem and our proposed solution. Three distinct clusters in Euclidean space are sampled in two trials (batches) and the eigenspace embedding is computed analytically. Three samples appear in both subsets, one for each cluster (red color). We refer to the common samples as \emph{anchors}. The plots of the instances in the embedding space for the two subsets are shown in Figs. \ref{fig:toyExmple_LS1}-\ref{fig:toyExmple_LS2}. One can observe the embeddings are different. Specifically, all anchors, which appear in both samplings, are mapped differently in a substantial way. It is well known that eigenvector embedding has a degree of freedom of rotation (as shown for example in  \cite{zelnik2004self}). However, in the case of uneven sampling of clusters there may be also some scale changes and slight translation (caused by the orthonormality constraints). We thus approximate these degrees of freedom in the general case as an affine rigid transformation according to the anchors. Aligning one embedding space to the other one, using this transformation, yields a consistent embedding, as can be seen in \cref{fig:toyExmple_LS2wthLS1}. Following the alignment process, the embedding can be learnt well using batches.

\begin{figure}[htbp!]
    \centering
    \begin{subfigure}[H]{0.23\textwidth}
        \centering
        \includegraphics[trim = 1mm 1mm 1mm 1mm, clip=true,width=1\textwidth]{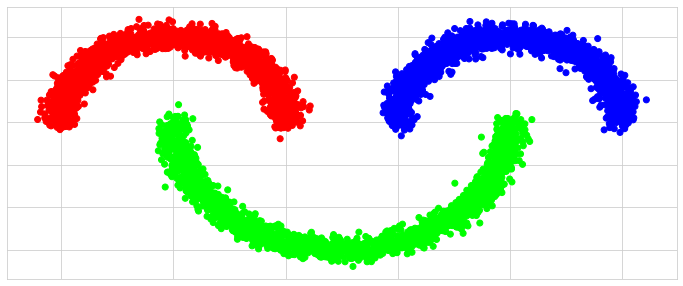}
        \caption{}
        \label{fig:3_moons_data_set_final}
    \end{subfigure}
    \hfill
    \begin{subfigure}[H]{0.23\textwidth}
        \centering
        \includegraphics[trim = 1mm 1mm 1mm 1mm, clip=true,width=1\textwidth]{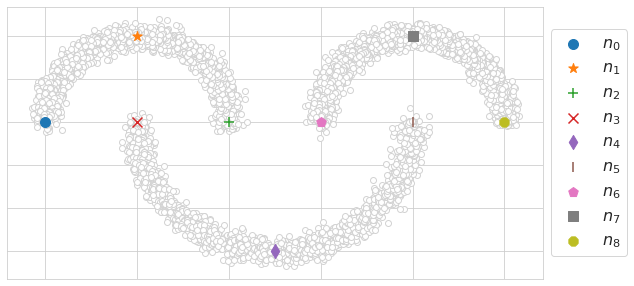}
        \caption{}
        \label{fig:3_moons_data_set_anchors}
    \end{subfigure}
    \hfill
    \begin{subfigure}[H]{0.23\textwidth}
        \centering
        \includegraphics[trim = 1mm 1mm 1mm 1mm, clip=true,width=1\textwidth]{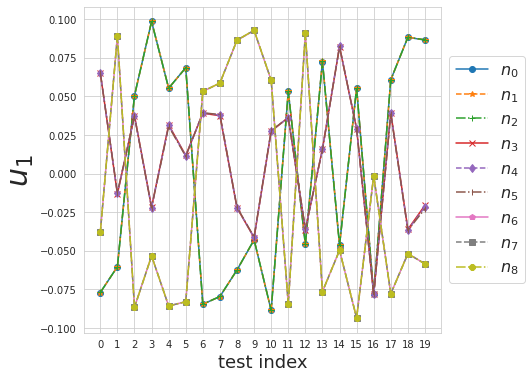}
        \caption{}
        \label{fig:3_moons_ev1_not_affined_final}
    \end{subfigure}
    \hfill
    \begin{subfigure}[H]{0.23\textwidth}
        \centering
        \includegraphics[trim = 1mm 1mm 1mm 1mm, clip=true,width=1\textwidth]{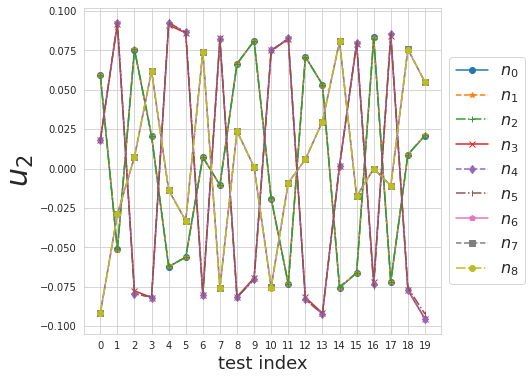}
        \caption{}
        \label{fig:3_moons_ev2_not_affined_final}
    \end{subfigure}
    \hfill
    \begin{subfigure}[H]{0.23\textwidth}
        \centering
        \includegraphics[trim = 1mm 1mm 1mm 1mm, clip=true,width=1\textwidth]{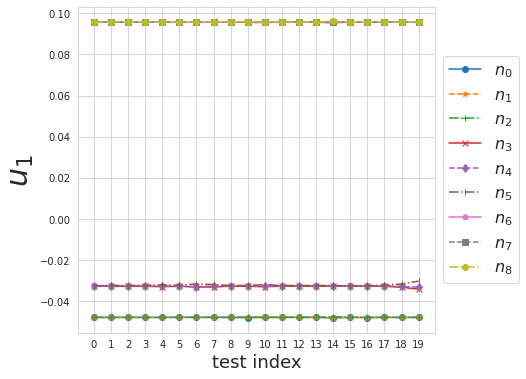}
        \caption{}
        \label{fig:3_moons_ev1_affined_final}
    \end{subfigure}
    \hfill
    \begin{subfigure}[H]{0.23\textwidth}
        \centering
        \includegraphics[trim = 1mm 1mm 1mm 1mm, clip=true,width=1\textwidth]{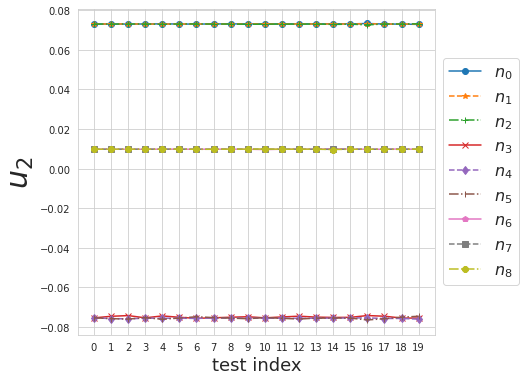}
        \caption{}
        \label{fig:3_moons_ev2_affined_final}
    \end{subfigure}
    \caption{{\bf Three-Moons toy example}. The full dataset is shown in \cref{fig:3_moons_data_set_final} and the chosen anchor-nodes in \cref{fig:3_moons_data_set_anchors}. Figs. \ref{fig:3_moons_ev1_not_affined_final}- \ref{fig:3_moons_ev2_not_affined_final} show the values of the two leading eigenvectors for the anchors, for 20 different graph-samples. Figs. \ref{fig:3_moons_ev1_affined_final}- \ref{fig:3_moons_ev2_affined_final} show those values after our proposed alignment.}
    \label{fig:3moons}
\end{figure}

In the toy example of the Three Moons, see \cref{fig:3moons}, we show the mapping of $9$ anchor-points, as shown in  \cref{fig:3_moons_data_set_anchors}. In Figs. \ref{fig:3_moons_ev1_not_affined_final}-\ref{fig:3_moons_ev2_not_affined_final} the analytic computation of the first two non-trivial eigenvectors are plotted for $20$ different batch samples of size $256$ (out of $9000$ nodes in the entire dataset), all of which contain the $9$ points. In this simple example anchors located on the same moon receive approximately the same value. However, in different batches the embedding of the anchors is clearly inconsistent. Surely, a network cannot be expected to generalize such a mapping. After learning the transformation and performing alignment, the embedding values are consistent. In Figs. \ref{fig:3_moons_ev1_affined_final}-\ref{fig:3_moons_ev2_affined_final} the values are shown after our correction procedure. 
This consistency allows to train DNN to learn the desired embedding, by dividing the data into batches.
The result of the trained DNN model for the Three Moons dataset appears in \cref{fig:model_output_all} (second from left).
These toy examples lead us to the detailed description of our algorithm. 
 
%


\subsection{{BASiS Algorithm}}
We propose to calculate the embedding space with batches. To obtain consistency in this representation, 
we calculate the first-order approximation of the distortion obtained in the eigenvector values between different samples of the data. The main steps of our algorithm are as follows: 
First we perform two preliminary initialization steps.

{\bf Defining an anchor set.} Draw $l$ samples from the data. This subset is denoted as $V^a$ and will be added to any batch in the learning process.

{\bf Defining the reference embedding space.} We would like to define the embedding space of the anchor set as a reference space. However, to get more information about the manifold $\mathcal{M}$, we add $m-l$ samples (randomly) and term it as the reference set $V^{ref}$. After calculating the embedding $V^{ref} \to \varphi^{ref}$ (as in \cref{embedding_def}), one can extract the coordinates of the anchor samples, 
\begin{equation}\label{eq:stepEmbeddingRef} 
    V^a\to \{\varphi^{a, ref}_i\}_{i=1}^l.
\end{equation}

Following this initialization, the main steps of the training are as follows:

{\bf Calculate the embedding space over a new batch.} Draw $m-l$ new samples and add them to the anchor set. Let us denote the union set as $V^{b}$. We calculate $\{\varphi_i\}_{i=1}^m$, the embedding of $V^{b}$ and extract the embedding $\{\varphi^{a}_i\}_{i=1}^l$  corresponding to the anchors. 

{\bf Calculate the alignment transformation.} Now, we calculate the alignment transformation between $\{\varphi^{a}_i\}_{i=1}^l$ to $\{\varphi^{a, ref}_i\}_{i=1}^l$. More formally, for $\varphi^a, \varphi^{a, ref}  \in \mathbb{R}^K$ we find $A \in \mathbb{R}^{K \times K}$  and $b \in \mathbb{R}^K$ which minimize
\begin{equation}\label{eq:affie_transformation_least_squares1}
        \min_{A, b}\sum_{i=1}^l{\norm{\varphi^{a, ref}_i - (A\varphi^{a}_i +b)}^2}.
\end{equation}
Alternatively, one can define $\hat{\varphi}^{a} = [\varphi^{a}, 1]$ and find the transformation $T \in \mathbb{R}^{K\times(K+1)}$ such that         \begin{equation}\label{eq:stepAffie_transformation_least_squares2}
            \min_{T}\sum_{i=1}^l{\norm{\varphi^{a, ref}_i - T\hat{\varphi}^{a}_i}^2}.
\end{equation}
In this case there are $K\times(K+1)$ degrees of freedom. Each anchor provides $K$ constraints, that means at least $K+1$ anchors are needed in order to solve this least squares problem. Since in many real-world problem there is noise in the measurements, it is customary to solve such problems in an overdertermined setting, using a higher number of anchors. In addition, given a large number of anchors, the transformation can be calculated using best matches - for example by using the RANdom SAmple Consensus (RANSAC) algorithm \cite{fischler1981random}.

{\bf Batch Alignment.} Given the transformation $T$, we can align the embedding $\{\varphi_i\}_{i=1}^m$ of all the instances of $V^{b}$. We  define $\hat{\varphi} = [\varphi, 1]$ and update
\begin{equation}\label{eq:stepAlignment}
     \varphi \leftarrow T\hat{\varphi}.
\end{equation}

{\bf Gradient Step.} Now that we have a mechanism that allows to get consistent embedding, we can train the DNN by dividing the data into batches and use a simple MSE lose function
\begin{equation}\label{ltsnet_loss}
    \mathcal{L}_{BASiS}(\theta)=\frac{1}{m}\sum_{i=1}^m\norm{y_i-\varphi_i}^2,
\end{equation}
where $y_i = f_{\theta}(x_i)$ is the DNN's output and $\varphi_i$ is the embedding of $x_i$ after alignment.

The full training scheme is detailed in \Cref{ltsnet_training_scheme}.




\begin{algorithm}[htbp!]
\caption{BASiS Training Scheme}
\label{ltsnet_training_scheme}
\begin{algorithmic}[1]
\Inputs{data features $\{x_i \}_{i=1}^n$, number of eigenvectors $K$, batch size $m$, number of anchors $l$.}
\Outputs{Spectral embedding model $f: \mathcal{M} \rightarrow \mathbb{R}^K $}
\Initialize{
Define anchor set $V^a$.\\
Extract $\{\varphi^{a, ref}_i\}_{i=1}^l$, the reference embedding of $V^a$ using \cref{eq:stepEmbeddingRef}
}
\While {$\mathcal{L}_{BASiS}(\theta)$ not converged}
 \State Draw $m-l$ new samples.
 \State Define node set $V^b$ as union of the anchors with the new sampled nodes.
 \State Calculate the embedding $\{\varphi_i\}_{i=1}^m$ of $V^b$. 
 \State Calculate the optimal transformation $T$, \cref{eq:stepAffie_transformation_least_squares2}.
 \State Align $\{\varphi_i\}_{i=1}^m$  with $T$ , \cref{eq:stepAlignment}. 
 \State Do a gradient step of $\mathcal{L}_{BASiS}$, \cref{ltsnet_loss}.
\EndWhile
\end{algorithmic}
\label{algo:TrainingScheme}
\end{algorithm}

\subsection{BASiS for feature perturbation}
In the process of network training, the features are often not fixed and slightly change each iteration during training. 
In this case the adjacency values change and hence naturally also the embedding space. We suggest an algorithm to allow iterative changes in the feature space (inducing different graph metric). This algorithm is also based on an alignment technique.
Similar to \Cref{algo:TrainingScheme}, we define anchor nodes. Given the current features, we extract $\{\varphi^{a, prev}_i\}_{i=1}^l$, the current spectral embedding of the anchors. When the features are updated, we find the new embedding of the anchors $\{\varphi^{a, update}_i\}_{i=1}^l$. In order to maintain consistency in the learning process, we find a transformation $T_G$, as in \cref{eq:stepAffie_transformation_least_squares2}, that aligns the updated anchor embedding to the previous one. Then we align the entire embedding space according to the calculated transformation. \Cref{algo:global_transformation_algo}
summarizes the proposed method.


\begin{algorithm}[htb]
\caption{BASiS under iterative feature change}
\begin{algorithmic}[1]
\Inputs{ $\{\varphi^{a, prev}_i\}_{i=1}^l$ the anchors embedding over previous features $\{x_i\}_{i=1}^n$, updated features $\{ \hat{x}_i \}_{i=1}^n$.}
\Outputs{$\{\varphi^{update}_i\}_{i=1}^n$ the spectral embedding over the updated features, aligned to  $\{\varphi^{a, prev}_i\}_{i=1}^l$.}
\State Calculate the embedding $\{\varphi^{update}_i\}_{i=1}^n$  over the updated features.
\State Extract $\{\varphi^{a, update}_i\}_{i=1}^l$ the embedding correspond to the anchors.
\State Calculate the transformation $T_G$ between $\{\varphi^{a, prev}_i\}_{i=1}^l$ and $\{\varphi^{a, update}_i\}_{i=1}^l$.
\State  Align $\{\varphi^{update}_i\}_{i=1}^n$  with $T_G$.
\end{algorithmic}
\label{algo:global_transformation_algo}
\end{algorithm}

\begin{table*}[htb]
\centering
  \begin{tabular}{P{2.0cm}||P{3.0cm}|| P{2.4cm} P{2.4cm} P{2.4cm} P{2.4cm}}
  \toprule
    
    Measures &  Networks & MNIST & Fashion-MNIST   & SVHN &  CIFAR-10  \\
    \midrule
    \multirow{3}{*}{$d_G$↓}        & Diffusion-Net  & $0.204 \pm 0.058$       & $0.488 \pm 0.238$      & $1.909 \pm 0.238$      & $1.022 \pm 0.250$ \\
                                   & SpecNet1       & $0.386 \pm 0.074$       & $0.375 \pm 0.132$      & $3.526 \pm 0.529$      & $2.256 \pm 0.471$ \\
                                   & SpecNet2       & $1.388 \pm 0.262$       & $1.976 \pm 0.210$      & $1.903 \pm 0.242$      & $2.970 \pm 0.682$ \\
                                   & BASiS (Ours)         & $\bf{0.107 \pm 0.038}$  & $\bf{0.284 \pm 0.073}$ & $\bf{1.656 \pm 0.170}$ & $\bf{0.803 \pm 0.085}$ \\
    \midrule
    \multirow{3}{*}{$d_{\perp}$↓}  & Diffusion-Net  & $0.535 \pm 0.365$       & $0.823 \pm 0.664$       & $1.532 \pm 0.354$      & $2.957 \pm 1.837$ \\
                                   & SpecNet1       & $6.296 \pm 0.922$       & $6.384 \pm 0.899$       & $4.507 \pm 0.821$      & $5.169 \pm 0.775$ \\
                                   & SpecNet2       & $9.486 \pm 0.001$       & $8.561 \pm 1.397$       & $4.104 \pm 0.269$      & $4.922 \pm 0.102$  \\
                                   & BASiS (Ours)   & $\bf{0.247 \pm 0.076}$       & $\bf{0.590 \pm 0.144}$       & $\bf{0.488 \pm 0.098}$      & $\bf{0.407 \pm 0.095}$ \\
    \midrule
    \multirow{3}{*}{NMI↑}          & Diffusion-Net  & $0.944 \pm 0.041$       & $0.759 \pm 0.085$      & $0.645 \pm 0.016$        & $0.466 \pm 0.034$ \\
                                   & SpecNet1       & $0.911 \pm 0.008$       & $0.761 \pm 0.011$      & $0.665 \pm 0.018$        & $0.443 \pm 0.012$ \\
                                   & SpecNet2       & $0.925 \pm 0.012$       & $0.759 \pm 0.010$      & $0.701 \pm 0.009$      & $0.466 \pm 0.013$ \\
                                   & BASiS (Ours)   & $\bf{0.961 \pm 0.001}$  & $\bf{0.798 \pm 0.001}$ & $\bf{0.736 \pm 0.001}$ & $\bf{0.501 \pm 0.001}$ \\
    \midrule
    \multirow{3}{*}{ACC↑}          & Diffusion-Net  & $0.944 \pm 0.030$       & $0.781 \pm 0.179$      & $0.687 \pm 0.303$        & $0.620 \pm 0.062$ \\
                                   & SpecNet1       & $0.963 \pm 0.005$       & $0.815 \pm 0.029$      & $0.811 \pm 0.039$        & $0.637 \pm 0.029$ \\
                                   & SpecNet2       & $0.966 \pm 0.007$       & $0.801 \pm 0.023$      & $0.813 \pm 0.015$      & $0.606 \pm 0.039$ \\
                                   & BASiS (Ours)   & $\bf{0.986 \pm 0.001}$  & $\bf{0.865 \pm 0.003}$ & $\bf{0.880 \pm 0.001}$ & $\bf{0.688 \pm 0.001}$ \\
    \midrule
    \multirow{3}{*}{Accuracy(\%)↑}     & Diffusion-Net  & $95.508 \pm 1.449$       & $86.207 \pm 0.196$      & $86.850 \pm 1.386$      & $67.316 \pm 2.112$ \\
                                   & SpecNet1       & $92.278 \pm 4.776$       & $84.123 \pm 1.229$      & $85.154 \pm 0.377$      & $65.336 \pm 0.626$ \\
                                   & SpecNet2       & $97.026 \pm 0.546$       & $85.953 \pm 0.240$      & $87.469 \pm 0.130$      & $67.093 \pm 0.644$ \\
                                   & BASiS (Ours)   & $\bf{98.522 \pm 0.065}$  & $\bf{87.202 \pm 0.187}$ & $\bf{88.021 \pm 0.064}$ & $\bf{68.887 \pm 0.128}$ \\
    \bottomrule
  \end{tabular}
  \caption{{\bf Spectral embedding performance comparison.} Average performance obtained over 10 different installations of each of the four methods. In each experiment we learn an embedding space in dimension of 10 for 1000 training iterations using batches of size 512.}
  \label{table:comparison_10ev_final}
\end{table*}

\section{Experimental Results}
In this section we examine the ability of BASiS to learn the graph-spectral embedding over different datasets quantifying its success using several performance measures. Our method is able to learn any desired eigen embedding (since it is supervised by analytic calculations). To fairly compare our method to the ones mentioned in \cref{sec:related_work} we calculate the eigenspace of $L_N$ (\cref{normalized laplacian}). For all methods the DNN's architecture includes $5$ fully connected (FC) layers with  ReLU activation in between (see full details in the supplementary). 

\subsection{Evaluation Metrics}
We evaluate our results using several measures. 
We calculate the Grassmann distance (projection metric) \cite{hamm2008grassmann} between the network output and the analytically calculated eigenvectors.
The squared Grassmann distance between two orthonormal matrices $Y_1, Y_2 \in \mathbb{R}^{n \times K} $ is defined as:
\begin{equation}\label{grassmann_distance_defenition}
d_G(Y_1,Y_2) = K-\sum_{i=1}^K{cos^2\theta_i},
\end{equation}
where $0 \leq \theta_1 \leq ... \leq \theta_K \leq \frac{\pi}{2}$ are the principal angles between the two subspaces $span(Y_1)$ and $span(Y_2)$. 
The distance is in $[0, K]$ where lower values indicate greater proximity between the subspaces.

A second measure is the degree of orthogonality of the DNN's output $Y$. 
We use the following orthogonality measure:
\begin{equation}\label{orthogonality_measure_defenition}
d_{\perp}(Y)=||Y^TY-I||_F^2,
\end{equation}
where $I$ is an identity  matrix and  $||\cdot||_F$ is the Frobenius norm. For $Y$ containing columns of orthonormal vectors we  get $d_{\perp}(Y)=0$. In general, we expect this measure to be close to zero in proper embeddings. 

To evaluate the potential clustering performance we examined two common metrics: Normalized mutual information (NMI) and unsupervised clustering accuracy (ACC). The clustering result is achieved by preforming the K-Means algorithm over the spectral embedding.  Both indicators are in the range $[0,1]$, where high values indicate a better correspondence between the clustering result and the true labels. 
NMI is defined as,
\begin{equation}\label{NMI_defenition}
NMI(c, \hat{c}) = \frac{I(c, \hat{c})}{max\{H(c), H(\hat{c})\}},
\end{equation}
where $I(c, \hat{c})$ is the mutual information between the true labels $c$ and the clustering result $\hat{c}$ and $H(\cdot)$ denotes entropy.  
ACC is defined as,
\begin{equation}\label{ACC_defenition}
ACC(c, \hat{c}) = \frac{1}{n}\max_{\pi \in \Pi}{\sum_{i=1}^n{\mathbbm{1}\{c_i=\pi(\hat{c}_i)\}}},
\end{equation}
where $\Pi$ is the set of possible permutations of the clustering results. To choose the optimal permutation $\pi$ we followed \cite{shaham2018spectralnet} and used the Kuhn-Munkres algorithm \cite{munkres1957algorithms}.

Finally, we examine how suitable the embedding is for classification. We trained (with Cross-Entropy loss) a supervised linear regression model containing a single fully connected layer without activation, and examined its accuracy:
\begin{equation}\label{accuracy_defenition}
Accuracy(c, \hat{c}) = \frac{1}{n}{\sum_{i=1}^n{\mathbbm{1}\{c_i=\hat{c}_i\}}}.
\end{equation}

\subsection{Spectral Clustering}
In this section we examine the ability of our method to learn the spectral embedding for clustering of different datasets. 
First, we examine the performance for well-known spectral clustering toy examples, appearing in \cref{fig:model_output_all}. In these examples the dataset includes $9000$ nodes and the model is trained by calculating the first non-trivial eigenvectors (sampling $256$ nodes, using $1000$ iterations). In all the experiments NMI and ACC of 1.0 were obtained over the test set (i.e., perfect clustering results). In addition, as demonstrate in \cref{fig:model_output_all}, the learnt model is able to generalize the clustering result and performs smooth interpolation and extrapolation of the space mapping. We note that no explicit regularization loss is used in our method, generalization and smoothness are obtained naturally through the neural training process.

Next we compare the performance of BASiS to those of the models mentioned in  \cref{sec:related_work}.
We examine the results over 4 well-known image datasets: MNIST, Fashion-MNIST \cite{xiao2017fashion}, SVHN \cite{netzer2011reading} and CIFAR-10 \cite{krizhevsky2009learning}. For each dataset we first learn an initial low-dimensional representation, found to be successful for graph-based learning, by a Siamese network, a convolutional neural network (CNN) trained in a supervised manner using Contrastive loss
\begin{equation}\label{contrastive_loss}
\begin{aligned}
\mathcal{L}_{Cont}(x_i, x_j, \theta) = 
\mathbbm{1}\{y_i=y_j\}\norm{f^{rep}_{\theta}(x_i)-f^{rep}_{\theta}(x_j)}_2^2\\ + \mathbbm{1}\{y_i \neq y_j\}\max(0, \epsilon-\norm{f^{rep}_{\theta}(x_i)-f^{rep}_{\theta}(x_j)}_2)^2,
\end{aligned}
\end{equation}
where $f^{rep}_{\theta}(x_i)$ is the Siamese network's output for input image $x_i$ labeld as $y_i$ , $\epsilon \in \mathbb{R}^+$.
More details on the architecture of the Siamese network are in the supplementary material.
We use this representations as inputs to the spectral embedding models. 
In all the experiments the graph affinity matrix $W$  is defined by \cref{Gaussian kernel}, using the $50$ nearest neighbors of each node. 
We use $50$ neighbors in order that all methods could perform well (see sensitivity analysis hereafter). 
The model is trained to learn the first $K$ eigenvectors, where $K$ is equal to the number of clusters. 
The batch size is set to $m=512$. For our method, we randomly select $25$ anchor-nodes from each cluster and use RANSAC to find the best transformation.

Comparison between the methods is summarized in \Cref{table:comparison_10ev_final}. The numbers are obtained by showing the average and empirical standard deviation of each measure, as computed based on $10$ experiments.   

Since the training process in Diffusion Net is not scalable, in each initialization we randomly sampled a single batch from the training set and trained the model with the analytically calculated spectral embedding. 
In relation to SpecNet2, as indicated in \cref{sec:related_work}, to obtain an approximation of the spectral space, SpecNet2 requires a post-processing stage over the network's output. In order to maintain consistency and obtain reasonable performance for all  measures, the post-processing is performed over the entire test set (this naturally limits the method and increases the level of complexity at inference). 
More implementation details are in the supplementary. \Cref{table:comparison_10ev_final} shows that our method is superior and more stable in  approximating the analytical embedding and in  clustering. 

\begin{figure}[htbp!]
\captionsetup[subfigure]{justification=centering}
    \centering
    \begin{subfigure}[t]{0.23\textwidth}
        \centering
        \includegraphics[trim = 1mm 1mm 1mm 1mm, clip=true,width=1\textwidth]{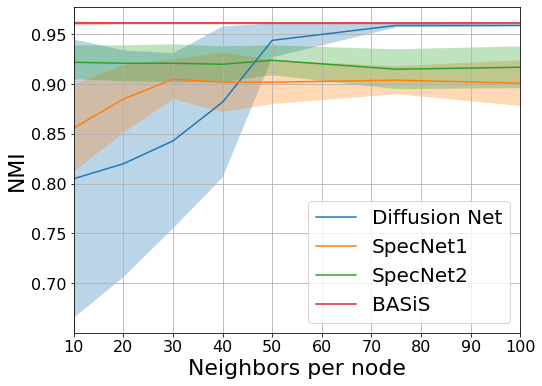}
        \caption{}
        \label{fig:ms_nmi}
    \end{subfigure}
    \hfill
    \begin{subfigure}[t]{0.23\textwidth}
        \centering
        \includegraphics[trim = 1mm 1mm 1mm 1mm, clip=true,width=1\textwidth]{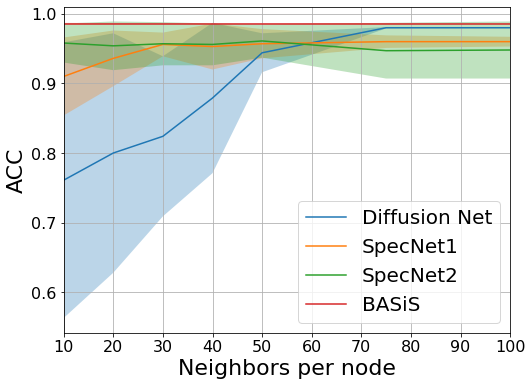}
        \caption{}
        \label{fig:ms_acc}
    \end{subfigure}
    \hfill
    \begin{subfigure}[t]{0.23\textwidth}
        \centering
        \includegraphics[trim = 1mm 1mm 1mm 1mm, clip=true,width=1\textwidth]{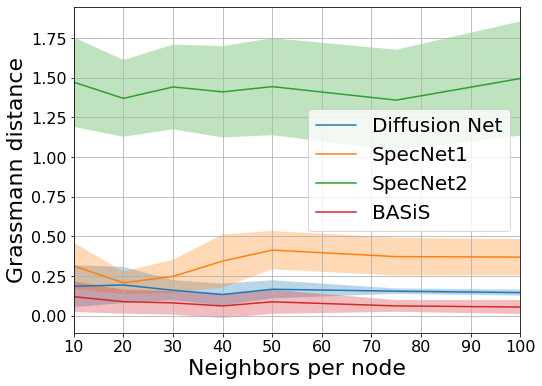}
        \caption{}
        \label{fig:ms_grassmann}
    \end{subfigure}
    \hfill
    \begin{subfigure}[t]{0.23\textwidth}
        \centering
        \includegraphics[trim = 1mm 1mm 1mm 1mm, clip=true,width=1\textwidth]{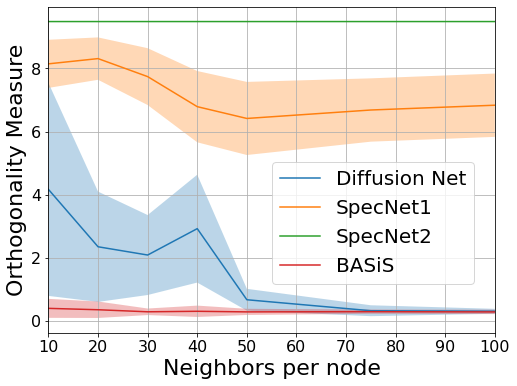}
        \caption{}
        \label{fig:ms_orth}
    \end{subfigure}
    \caption{{\bf Robustness to the node neighborhood.} The average and standard deviation of $50$ different training experiments over the MNIST dataset, for different number of neighbors per node.}
    \label{fig:ms_change}
\end{figure}

We further examined the robustness of the methods to changes in the number of neighbors for each node. This parameter defines the connectivity of the graph in the affinity matrix. \cref{fig:ms_change} shows the average and the empirical standard deviation of the  performance measures, for $50$ training procedures over the MNIST dataset.
It is shown that our method is highly robust and consistent. We note the high sensitivity of Diffusion Net to this meta-parameter.

\begin{figure}[htb]
\captionsetup[subfigure]{justification=centering}
    \centering
    \begin{subfigure}[t]{0.11\textwidth}
        \centering
        \includegraphics[trim = 1mm 1mm 1mm 1mm, clip=true,width=1\textwidth]{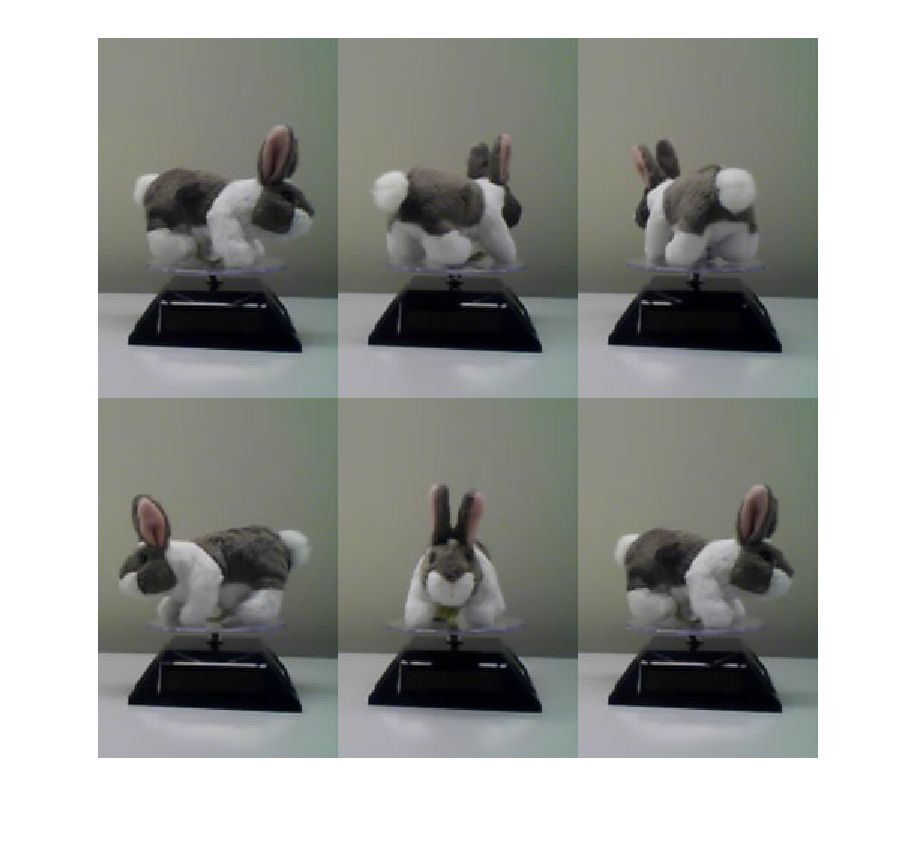}
        \caption{}
        \label{fig:bunny_image}
    \end{subfigure}
    \hfill
    \begin{subfigure}[t]{0.11\textwidth}
        \centering
        \includegraphics[trim = 1mm 1mm 1mm 1mm, clip=true,width=1\textwidth]{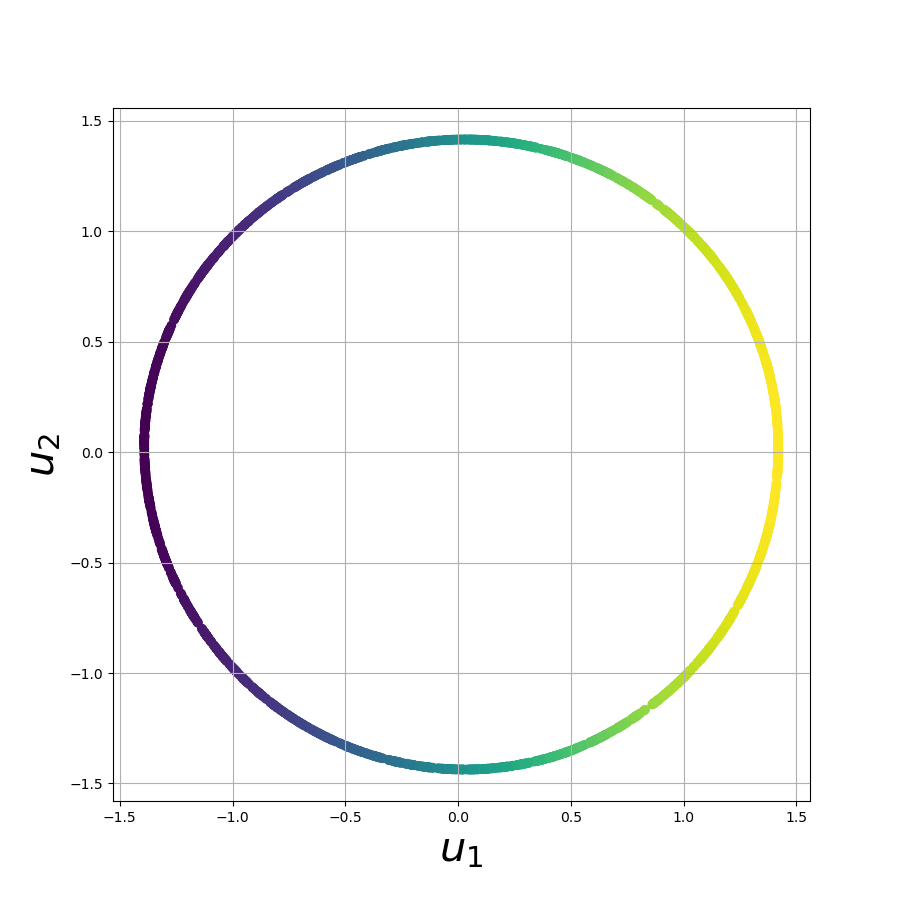}
        \caption{}
        \label{fig:full_data_dm}
    \end{subfigure}
    \hfill
    \begin{subfigure}[t]{0.11\textwidth}
        \centering
        \includegraphics[trim = 1mm 1mm 1mm 1mm, clip=true,width=1\textwidth]{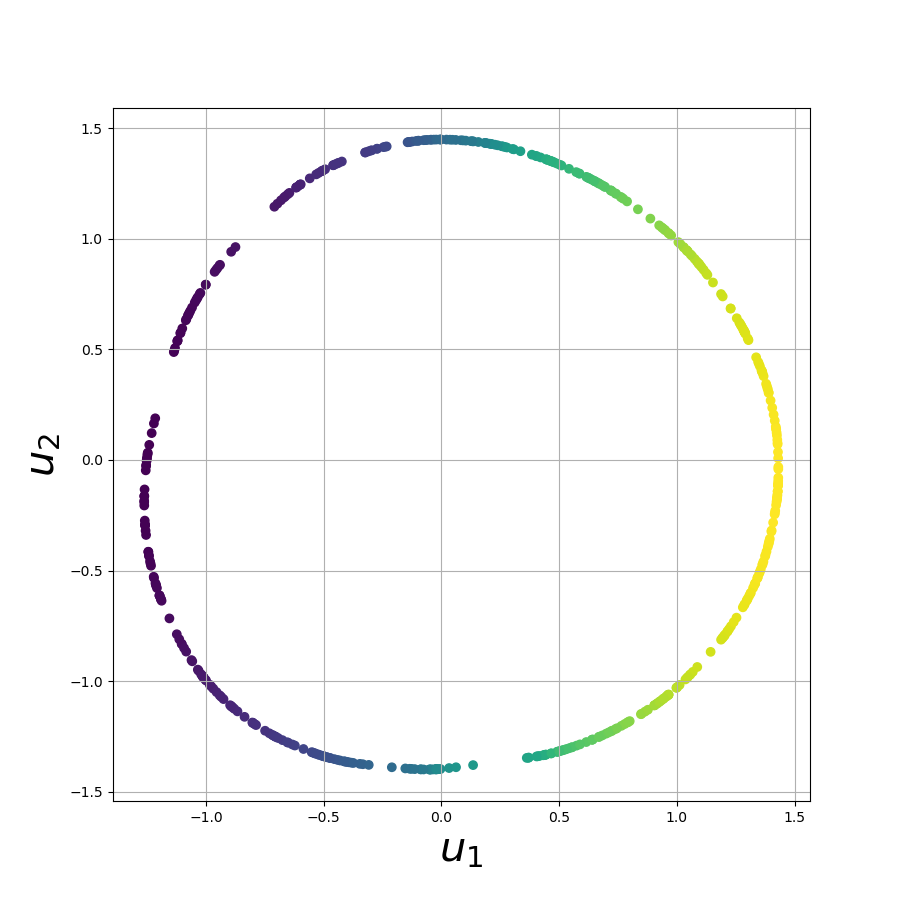}
        \caption{}
        \label{fig:test_set_dm}
    \end{subfigure}
    \hfill
    \begin{subfigure}[t]{0.11\textwidth}
        \centering
        \includegraphics[trim = 1mm 1mm 1mm 1mm, clip=true,width=1\textwidth]{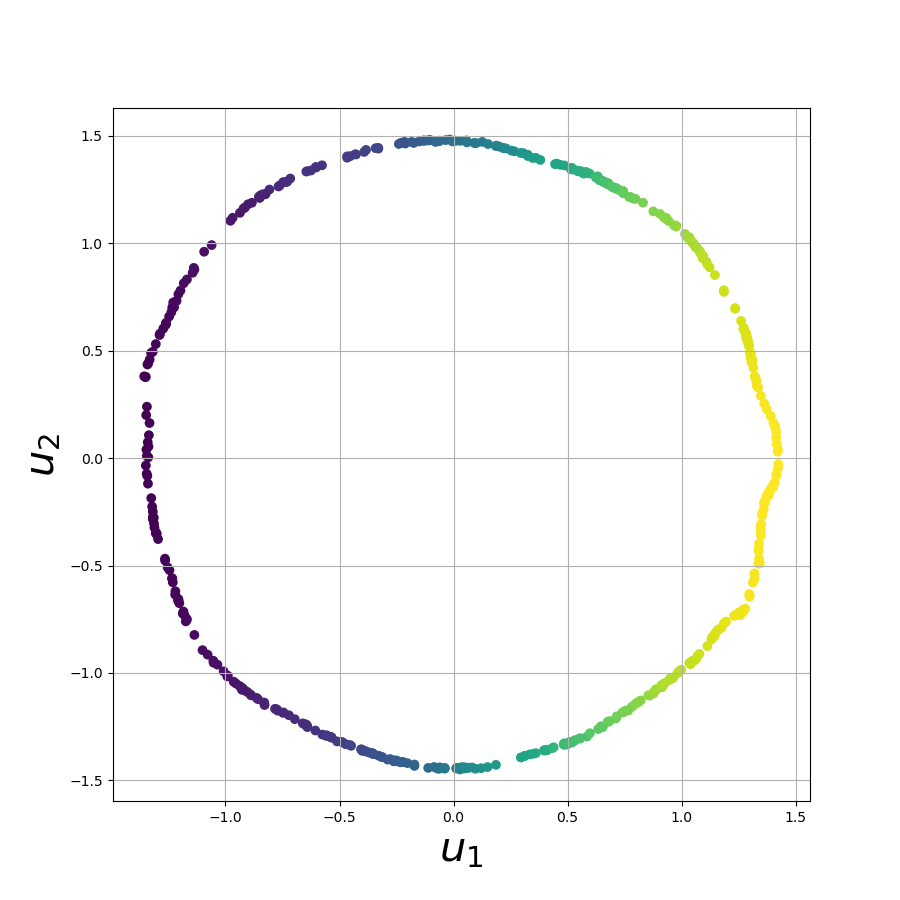}
        \caption{}
        \label{fig:test_set_net_output}
    \end{subfigure}
    \caption{{\bf Diffusion Maps encoding}. Data set of 2000 snapshots of bunny on rotating display \cite{lederman2018learning}. \cref{fig:bunny_image} shows snapshot examples. \cref{fig:full_data_dm} present the analytically calculated DM embedding for the full dataset . The test set analytical embedding is shown in \cref{fig:test_set_dm}  and the network output for the test set in \cref{fig:test_set_net_output}.}
    \label{fig:bunny_dm}
\end{figure}

\subsection{Diffusion Maps Encoding}
We examine the ability of our model to learn the DM embedding (\ref{dm_embedding_def}). For this purpose we use the dataset from \cite{lederman2018learning} which includes $2000$ snapshots of toy bunny located on a rotating surface. Six examples of such frames are shown in \cref{fig:bunny_image}. We define a graph using the $20$ nearest neighbors of each node, and calculate the random-walk matrix $P$ (\ref{eq:random_walk_matrix}). Raw pixels are used as features (dimension $288,000$). Dimension reduction is performed with DM to $\mathbb{R}^2$. In \cref{fig:full_data_dm} the analytically calculated embedding obtained based on the entire dataset is shown. Our model was trained to approximate this embedding using $1500$ images. Test is performed over $500$ images.
\cref{fig:test_set_dm} shows the analytically calculated embedding over the test snapshots. \cref{fig:test_set_net_output} shows the embedding obtained by our trained model. Our method approximate well the analytic calculation.

\subsection{Iterative Change of Features}
\begin{figure}[htb]
\captionsetup[subfigure]{justification=centering}
    \centering
    \begin{subfigure}[t]{0.20\textwidth}
        \centering
        \includegraphics[trim = 1mm 1mm 1mm 1mm, clip=true,width=1\textwidth]{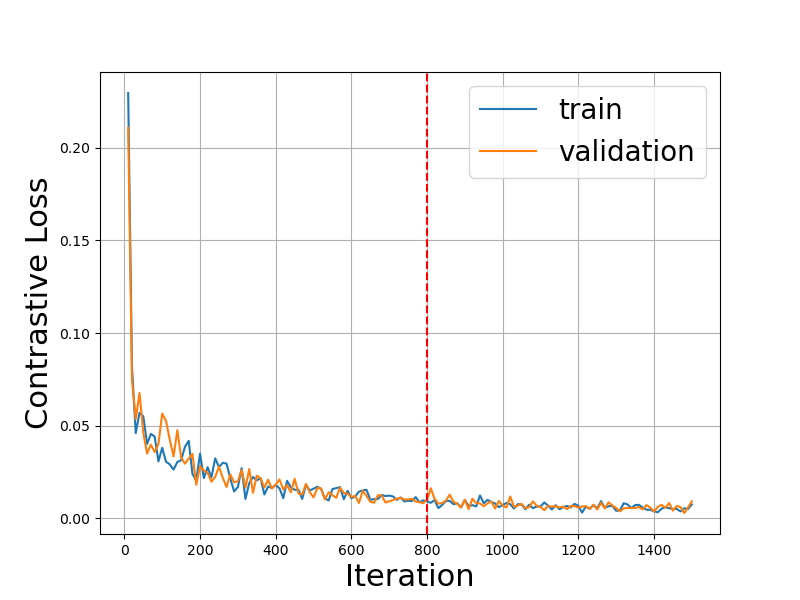}
        \caption{}
        \label{fig:tg_contrastive_loss}
    \end{subfigure}
    \begin{subfigure}[t]{0.20\textwidth}
        \centering
        \includegraphics[trim = 1mm 1mm 1mm 1mm, clip=true,width=1\textwidth]{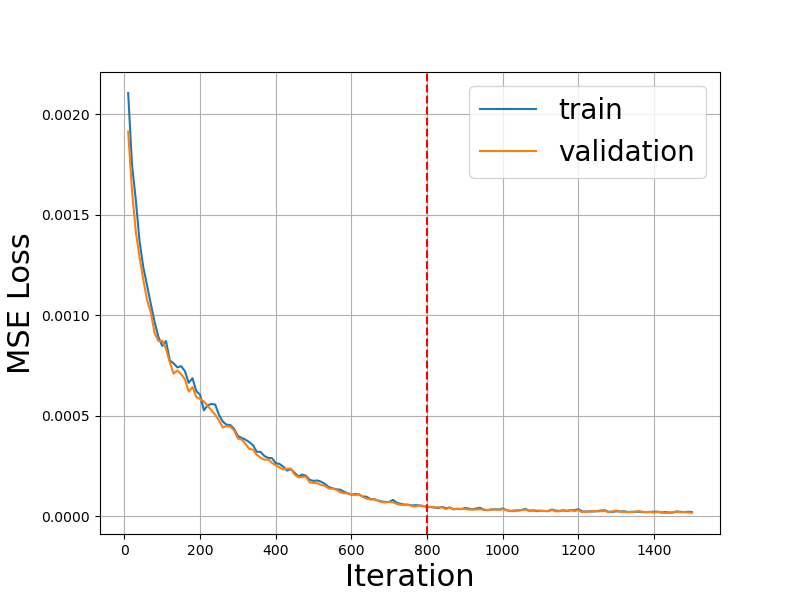}
        \caption{}
        \label{fig:tg_mse}
    \end{subfigure}
    \begin{subfigure}[t]{0.20\textwidth}
        \centering
        \includegraphics[trim = 1mm 1mm 1mm 1mm, clip=true,width=1\textwidth]{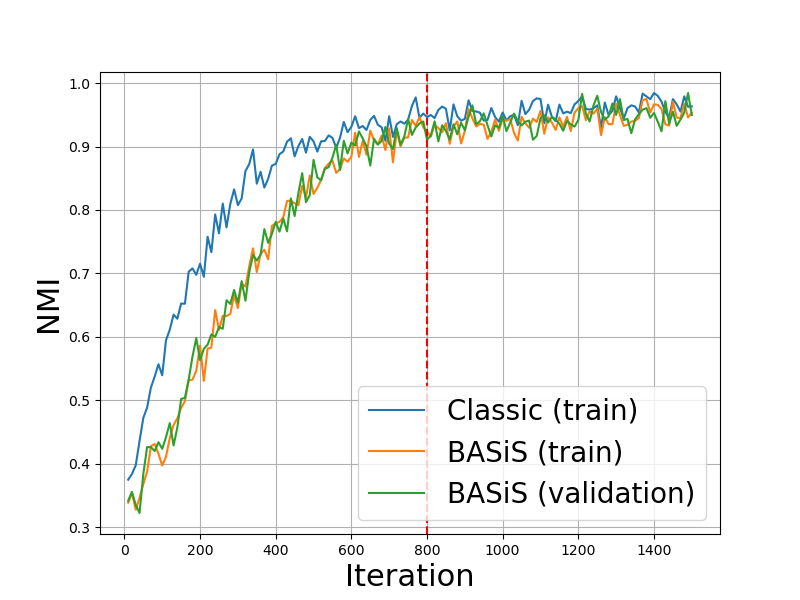}
        \caption{}
        \label{fig:tg_nmi}
    \end{subfigure}
    \begin{subfigure}[t]{0.20\textwidth}
        \centering
        \includegraphics[trim = 1mm 1mm 1mm 1mm, clip=true,width=1\textwidth]{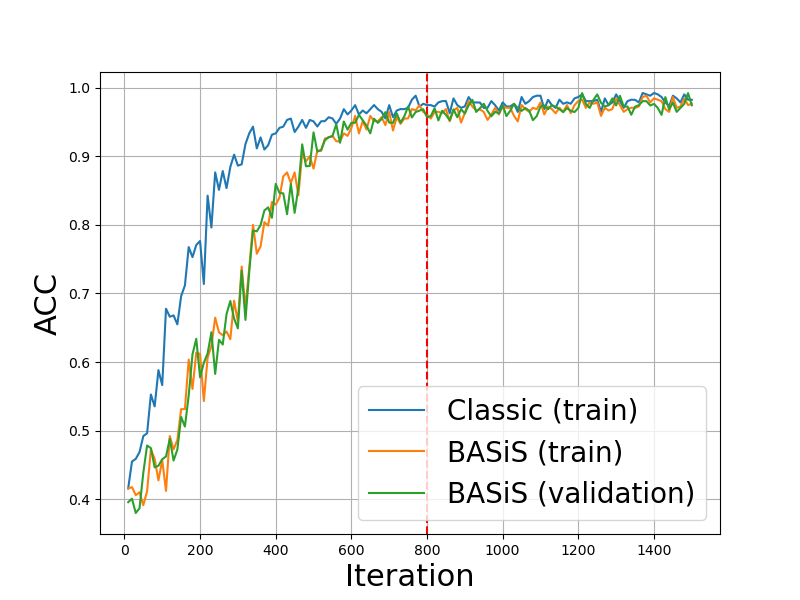}
        \caption{}
        \label{fig:tg_acc}
    \end{subfigure}
    \caption{{\bf Training under feature change}. Evolution of measures during training (MNIST). 
    \ref{fig:tg_contrastive_loss}-\ref{fig:tg_mse} losses of the features module and the spectral embedding module, respectively.  \ref{fig:tg_nmi}-\ref{fig:tg_acc} -- clustering measures. Blue, analytic calculation of the eigenvectors based on current features. Orange and green, output of spectral embedding module (training and validation sets, respectively).
  }
    \label{fig:tg_mnist_training}
\end{figure}

\begin{figure}[htb]
\captionsetup[subfigure]{justification=centering}
    \centering
    \begin{subfigure}[t]{0.23\textwidth}
        \centering
        \includegraphics[trim = 40mm 15mm 70mm 20mm, clip=true,width=1\textwidth]{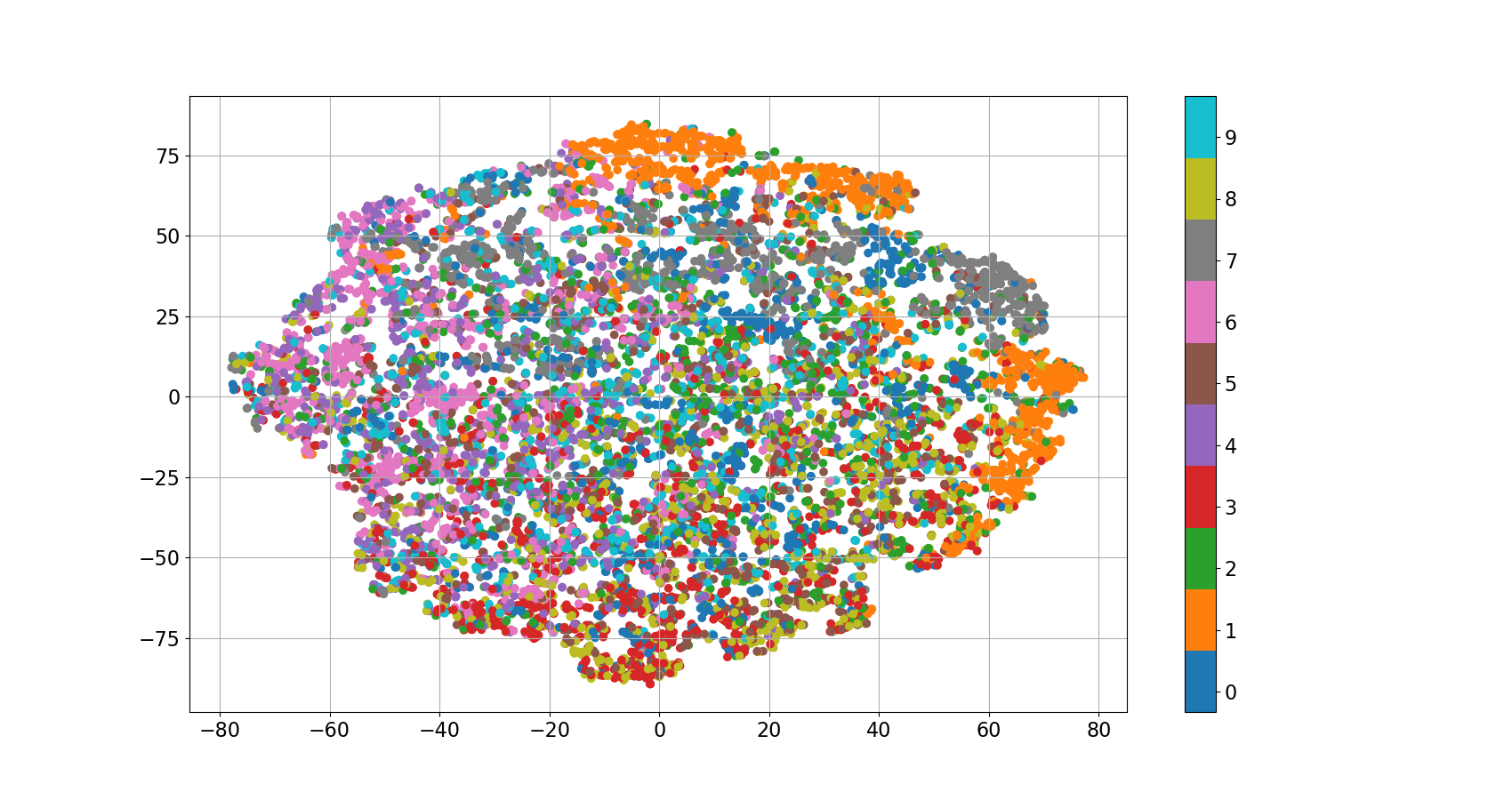}
        \caption{}
        \label{fig:ev_valid_iter1_tsne}
    \end{subfigure}
    \hfill
    \begin{subfigure}[t]{0.23\textwidth}
        \centering
        \includegraphics[trim = 40mm 15mm 70mm 20mm, clip=true,width=1\textwidth]{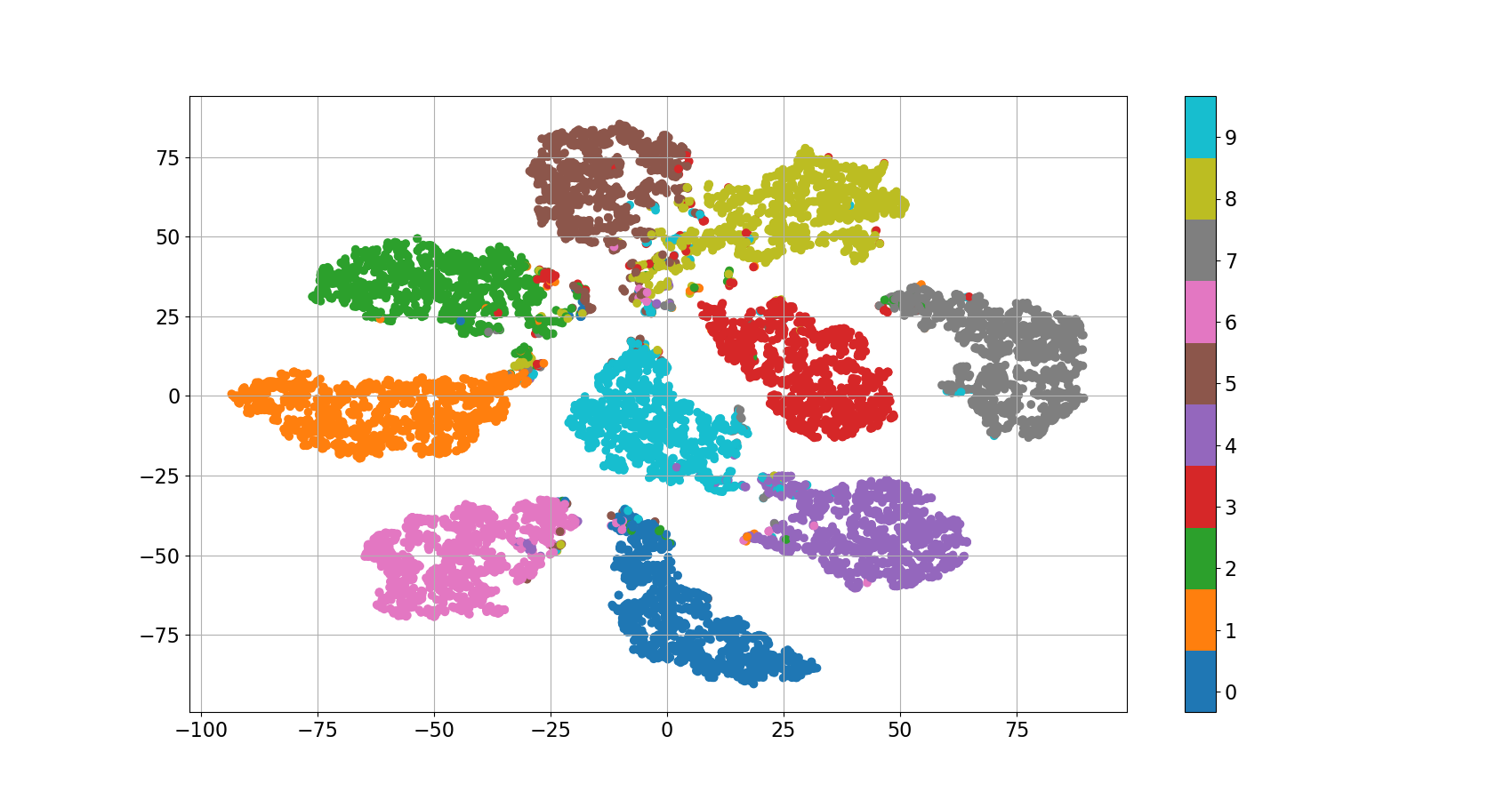}
        \caption{}
        \label{fig:ev_valid_iter1500_tsne}
    \end{subfigure}
    \caption{{\bf Test set embedding}. Visualization (TSNE) of the  spectral embedding, MNIST test set. 
    \cref{fig:ev_valid_iter1_tsne} shows the spectral embedding at the beginning of training,  \cref{fig:ev_valid_iter1500_tsne} at the end.}
    \label{fig:tg_mnist_embedding}
\end{figure}

In this section we illustrate the performance of \Cref{algo:global_transformation_algo} for aligning the spectral embedding space under changing features. 
We define two DNN models. The first one is for feature extraction, trained to minimize the Contrastive loss (\ref{contrastive_loss}). The second is trained for calculating the spectral embedding, using \Cref{algo:TrainingScheme}, based on the output of the features model.
Both models are trained simultaneously. The feature model is trained for $1500$ iterations, where every tenth iteration we perform an update step for the spectral embedding model. 
To maintain consistency under the feature change, we align the spectral space using \Cref{algo:global_transformation_algo} before performing an update step for the spectral model.
\cref{fig:tg_mnist_training} shows the results of the training process over the MNIST dataset where the learnt features are of dimension $16$ and the eigenvectors are of dimension $9$. \cref{fig:tg_mnist_embedding} shows a visualization (TSNE \cite{van2008visualizing}) 
of the test set embedding at the beginning and the end of the training process. The two modules were able to converge and to reach good  clustering results. In addition, when the loss of the spectral module is sufficiently low (around iteration 800, marked with a red line in \cref{fig:tg_mnist_training}) the clustering performance of the spectral module is comparable to the analytic embedding, calculated with the current features (the orange and the green plots tightly follow the blue plot).

\begin{figure}[htb]
\captionsetup[subfigure]{justification=centering}
    \centering
    \begin{subfigure}[t]{0.15\textwidth}
        \centering
        \includegraphics[trim = 40mm 15mm 70mm 20mm, clip=true,width=1\textwidth]{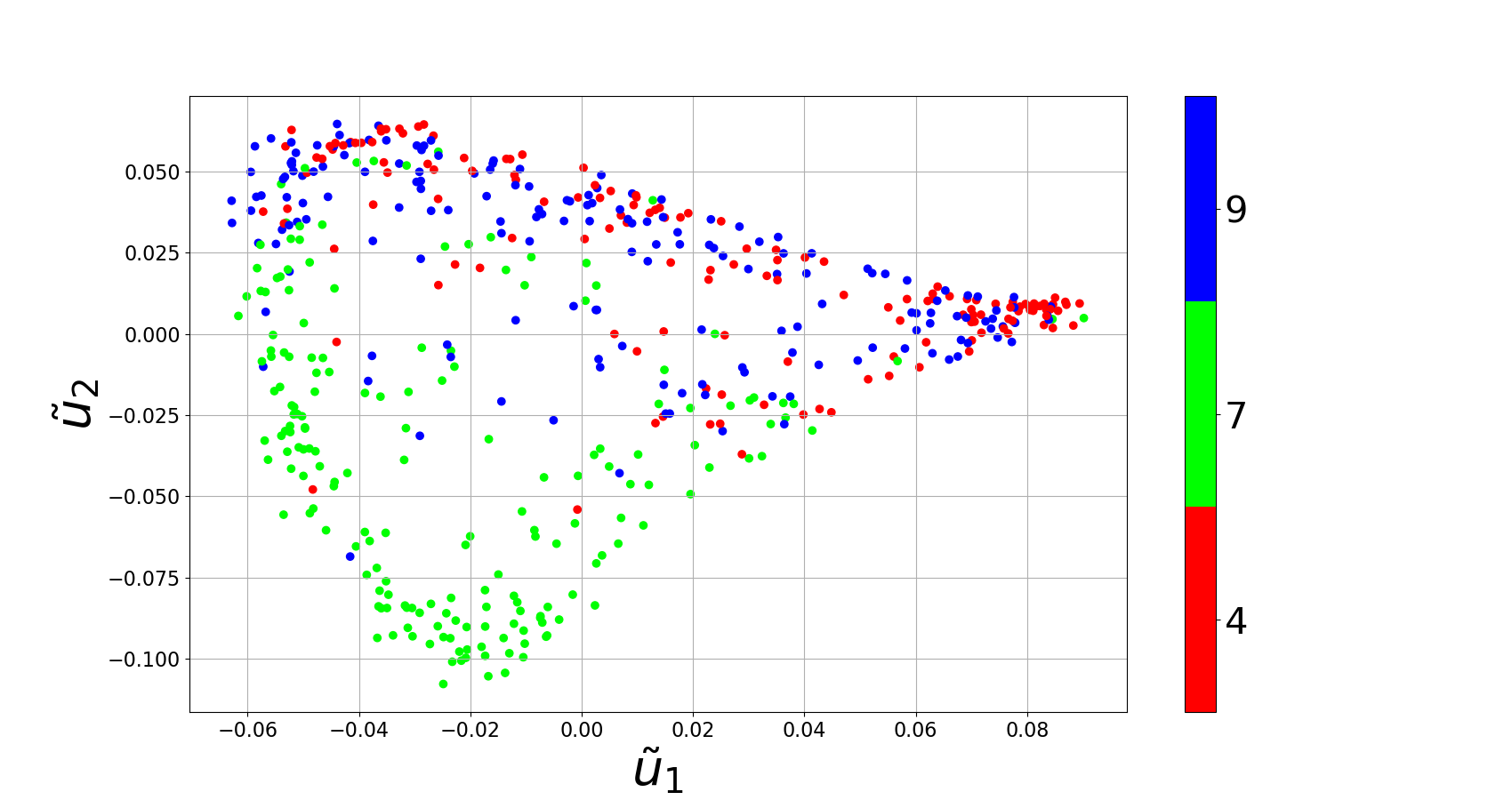}
        \label{fig:tg479_U_sampled_base_before_Tg_iter10}
    \end{subfigure}
    \hfill
    \begin{subfigure}[t]{0.15\textwidth}
        \centering
        \includegraphics[trim = 40mm 15mm 70mm 20mm, clip=true,width=1\textwidth]{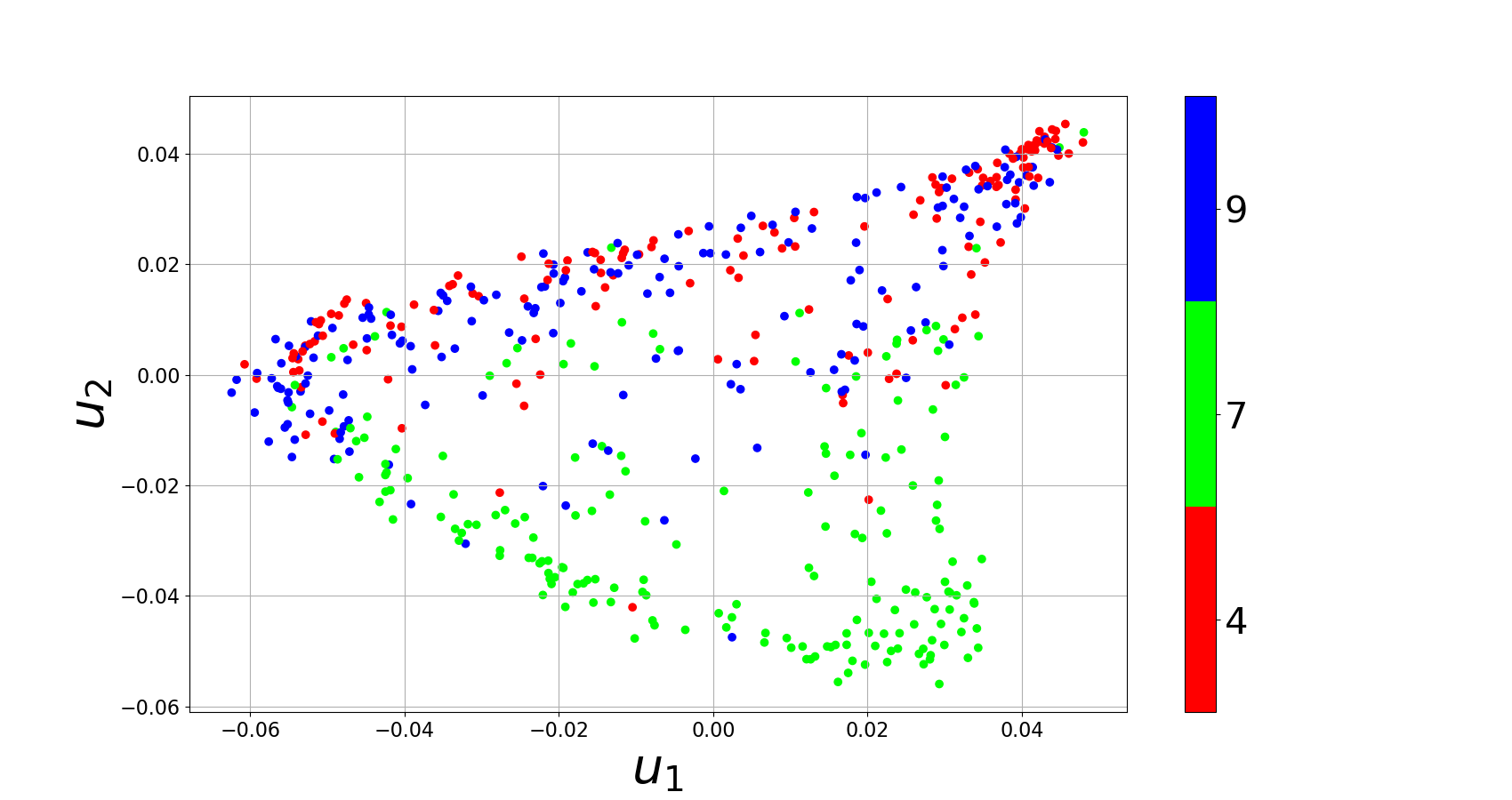}
        \label{fig:tg479_U_sampled_base_after_Tg_iter10}
    \end{subfigure}
    \hfill
    \begin{subfigure}[t]{0.15\textwidth}
        \centering
        \includegraphics[trim = 40mm 15mm 70mm 20mm, clip=true,width=1\textwidth]{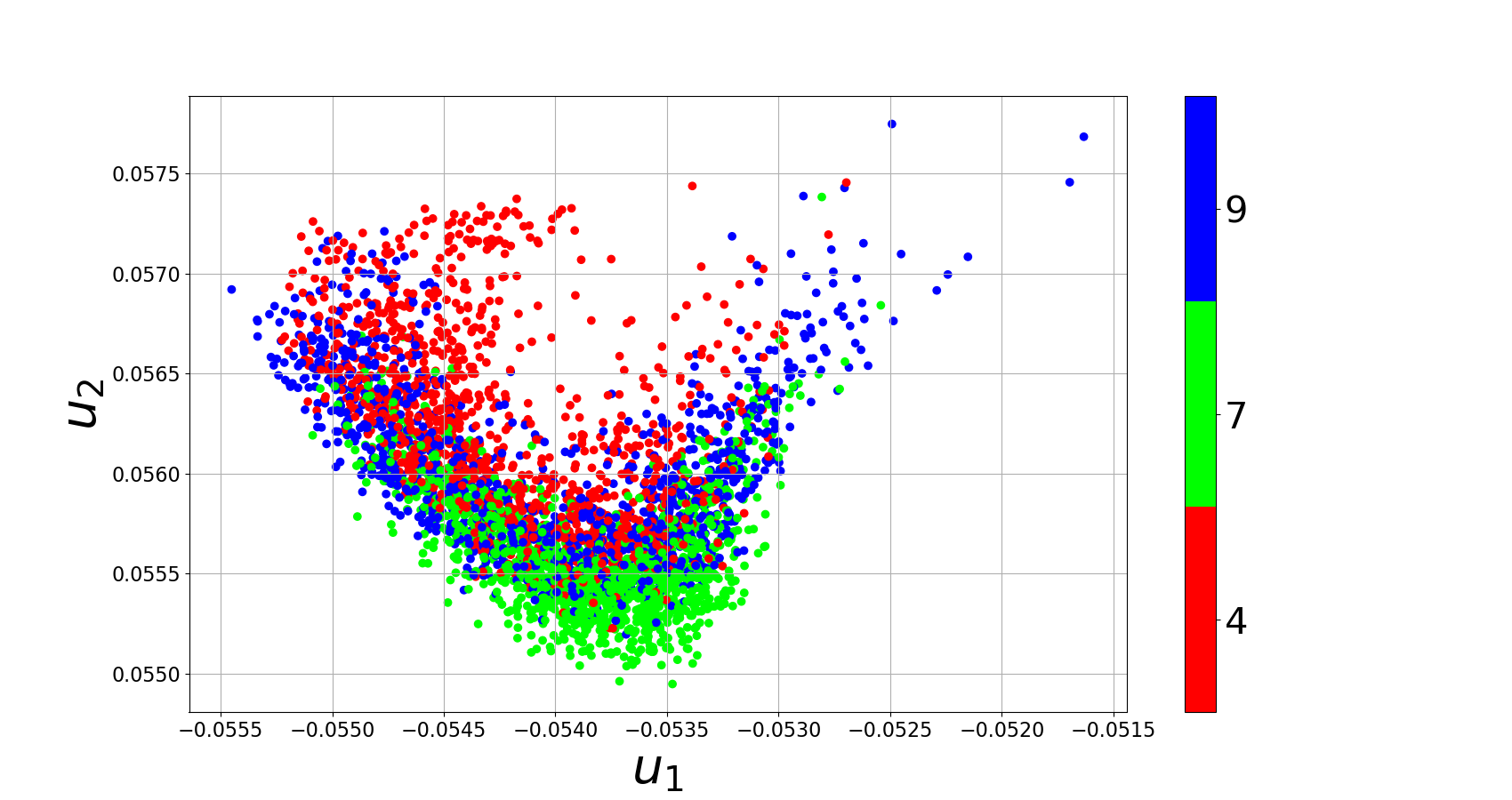}
        \label{fig:tg479_ev_valid_iter10}
    \end{subfigure}
    \hfill
    \begin{subfigure}[t]{0.15\textwidth}
        \centering
        \includegraphics[trim = 40mm 15mm 70mm 20mm, clip=true,width=1\textwidth]{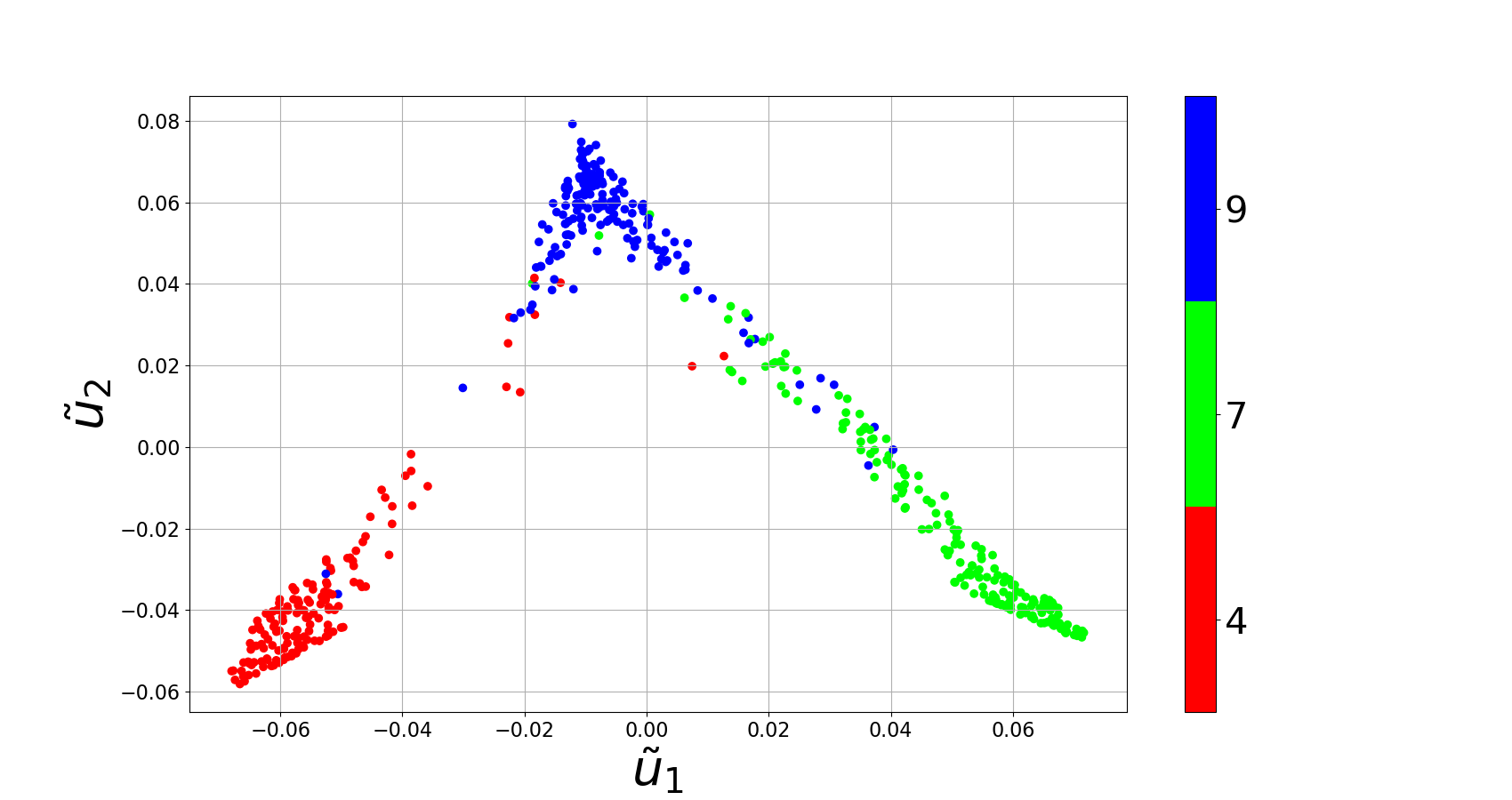}
        \label{fig:tg479_U_sampled_base_before_Tg_iter100}
    \end{subfigure}
    \hfill
    \begin{subfigure}[t]{0.15\textwidth}
        \centering
        \includegraphics[trim = 40mm 15mm 70mm 20mm, clip=true,width=1\textwidth]{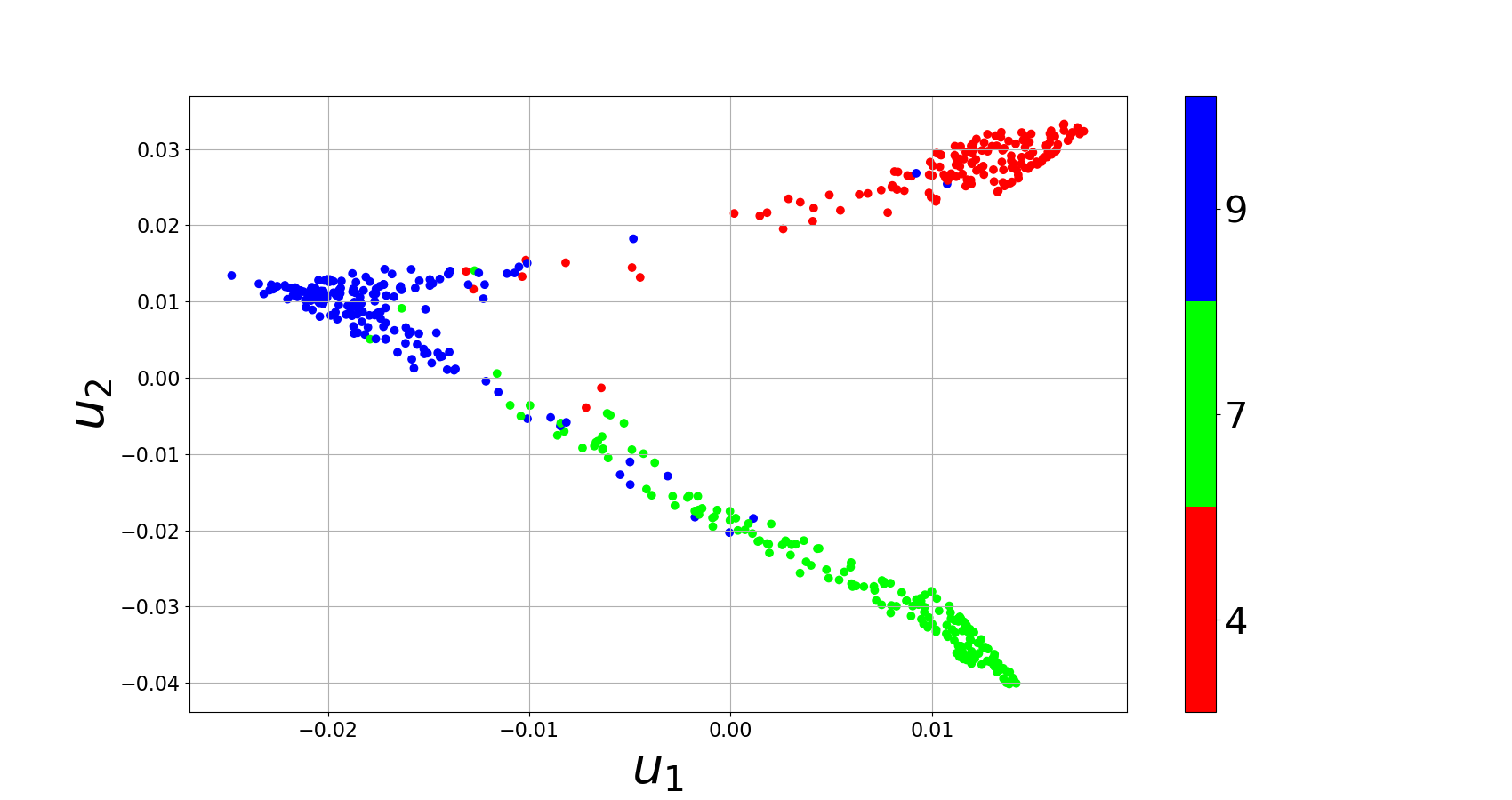}
        \label{fig:tg479_U_sampled_base_after_Tg_iter100}
    \end{subfigure}
    \hfill
    \begin{subfigure}[t]{0.15\textwidth}
        \centering
        \includegraphics[trim = 40mm 15mm 70mm 20mm, clip=true,width=1\textwidth]{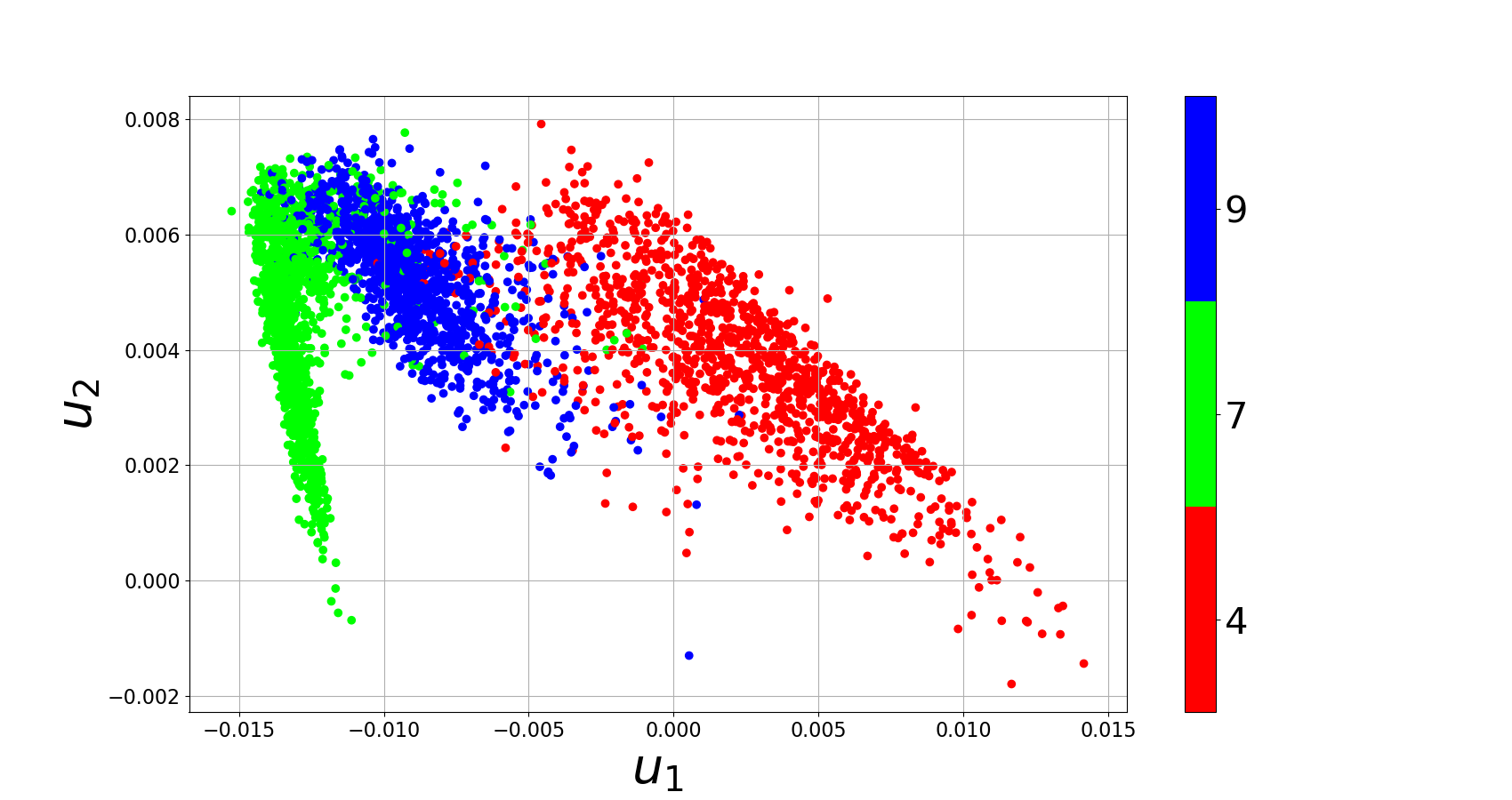}
        \label{fig:tg479_ev_valid_iter100}
    \end{subfigure}
    \hfill
    \begin{subfigure}[t]{0.15\textwidth}
        \centering
        \includegraphics[trim = 40mm 15mm 70mm 20mm, clip=true,width=1\textwidth]{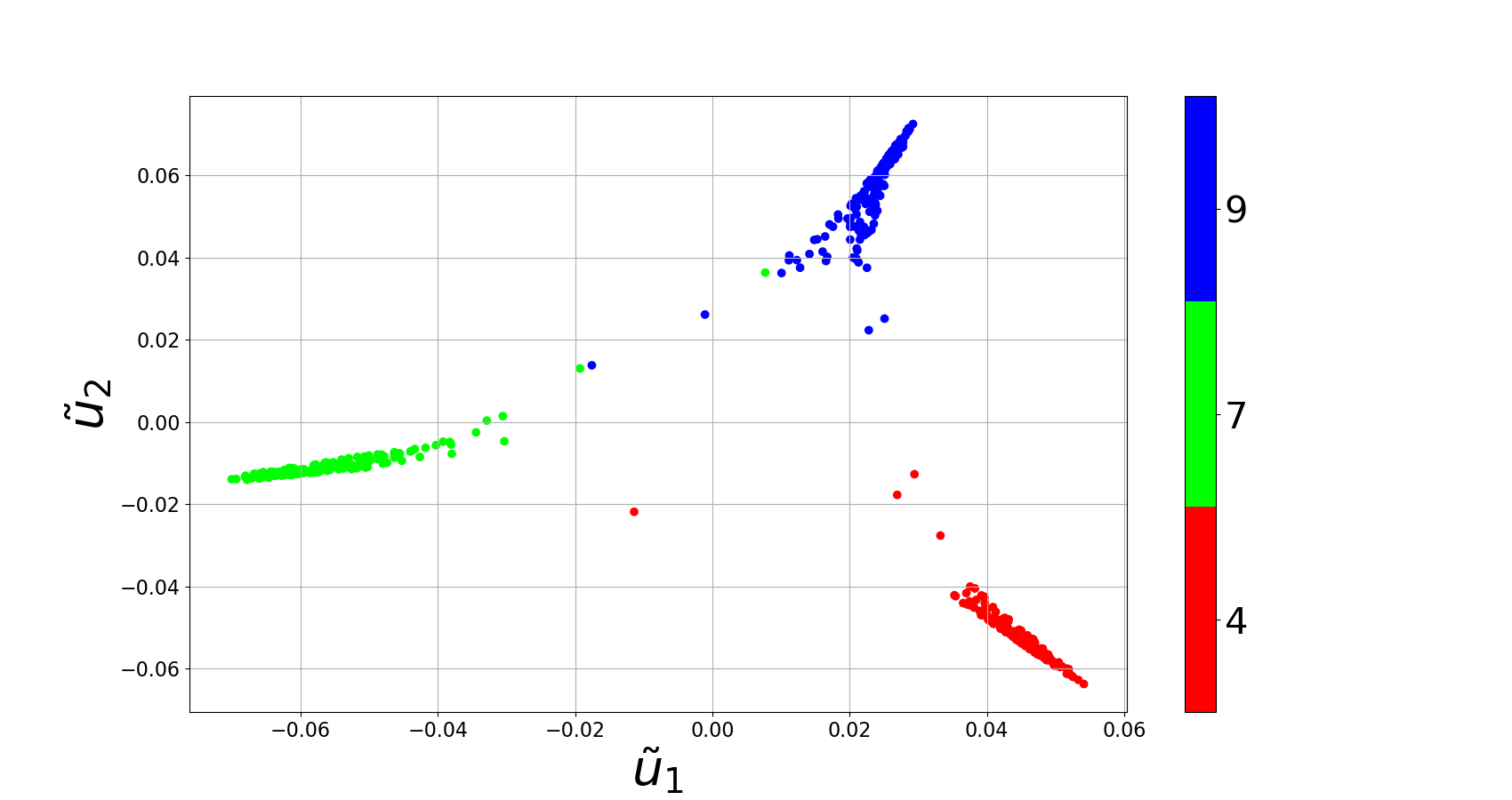}
        \label{fig:tg479_U_sampled_base_before_Tg_iter500}
    \end{subfigure}
    \hfill
    \begin{subfigure}[t]{0.15\textwidth}
        \centering
        \includegraphics[trim = 40mm 15mm 70mm 20mm, clip=true,width=1\textwidth]{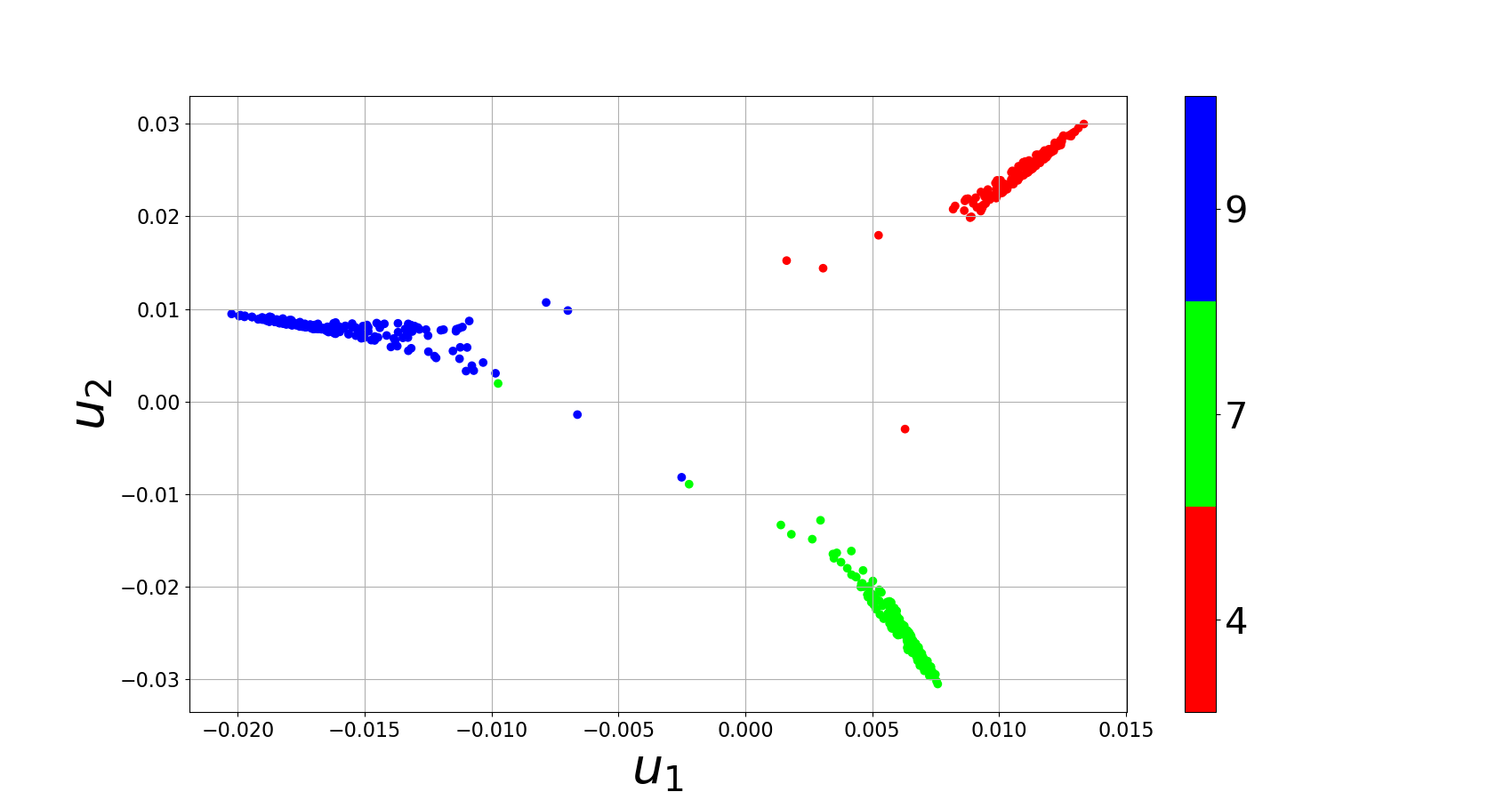}
        \label{fig:tg479_U_sampled_base_after_Tg_iter500}
    \end{subfigure}
    \hfill
    \begin{subfigure}[t]{0.15\textwidth}
        \centering
        \includegraphics[trim = 40mm 15mm 70mm 20mm, clip=true,width=1\textwidth]{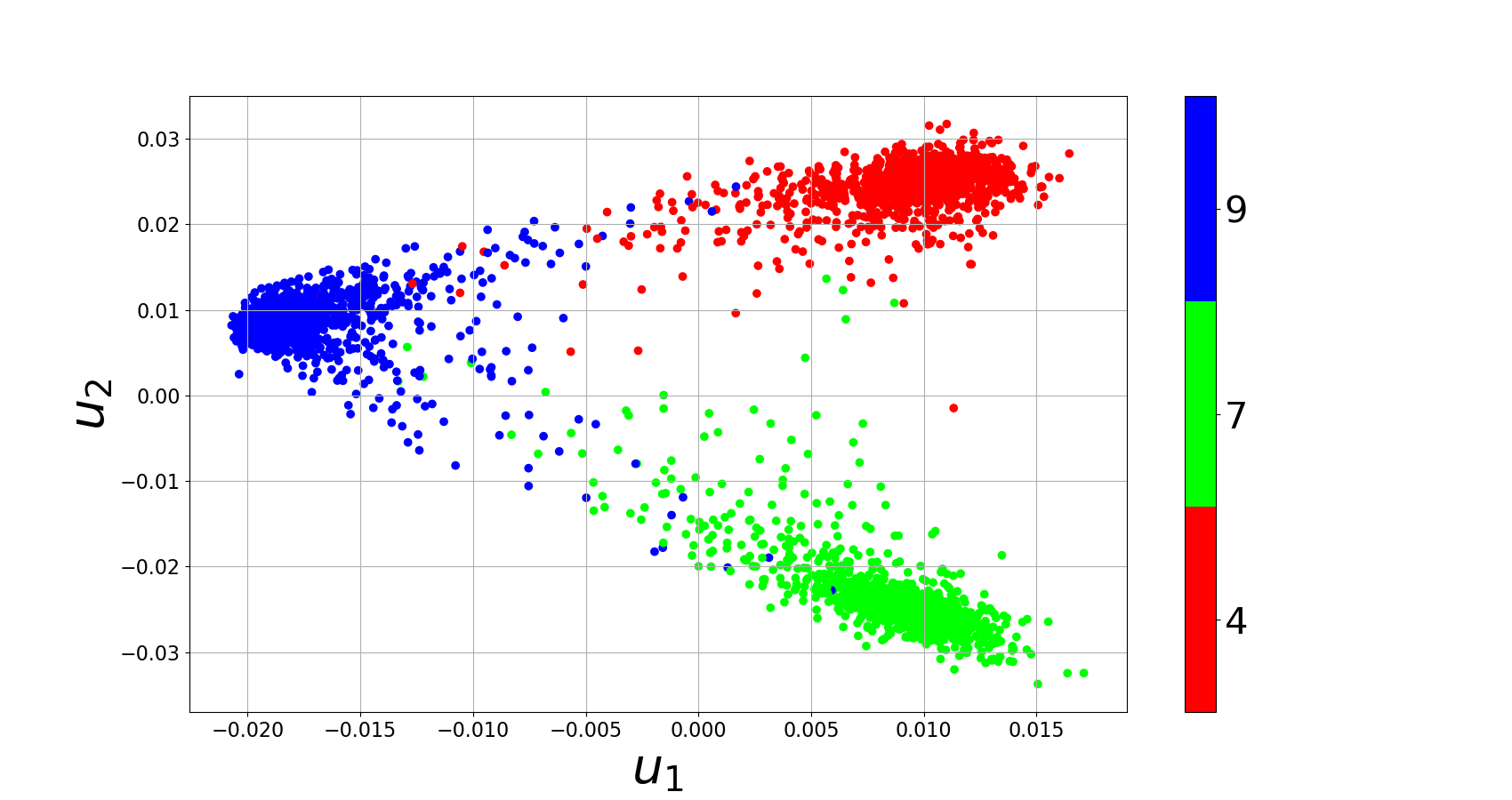}
        \label{fig:tg479_ev_valid_iter500}
    \end{subfigure}
    \hfill
    \begin{subfigure}[t]{0.15\textwidth}
        \centering
        \includegraphics[trim = 40mm 15mm 70mm 20mm, clip=true,width=1\textwidth]{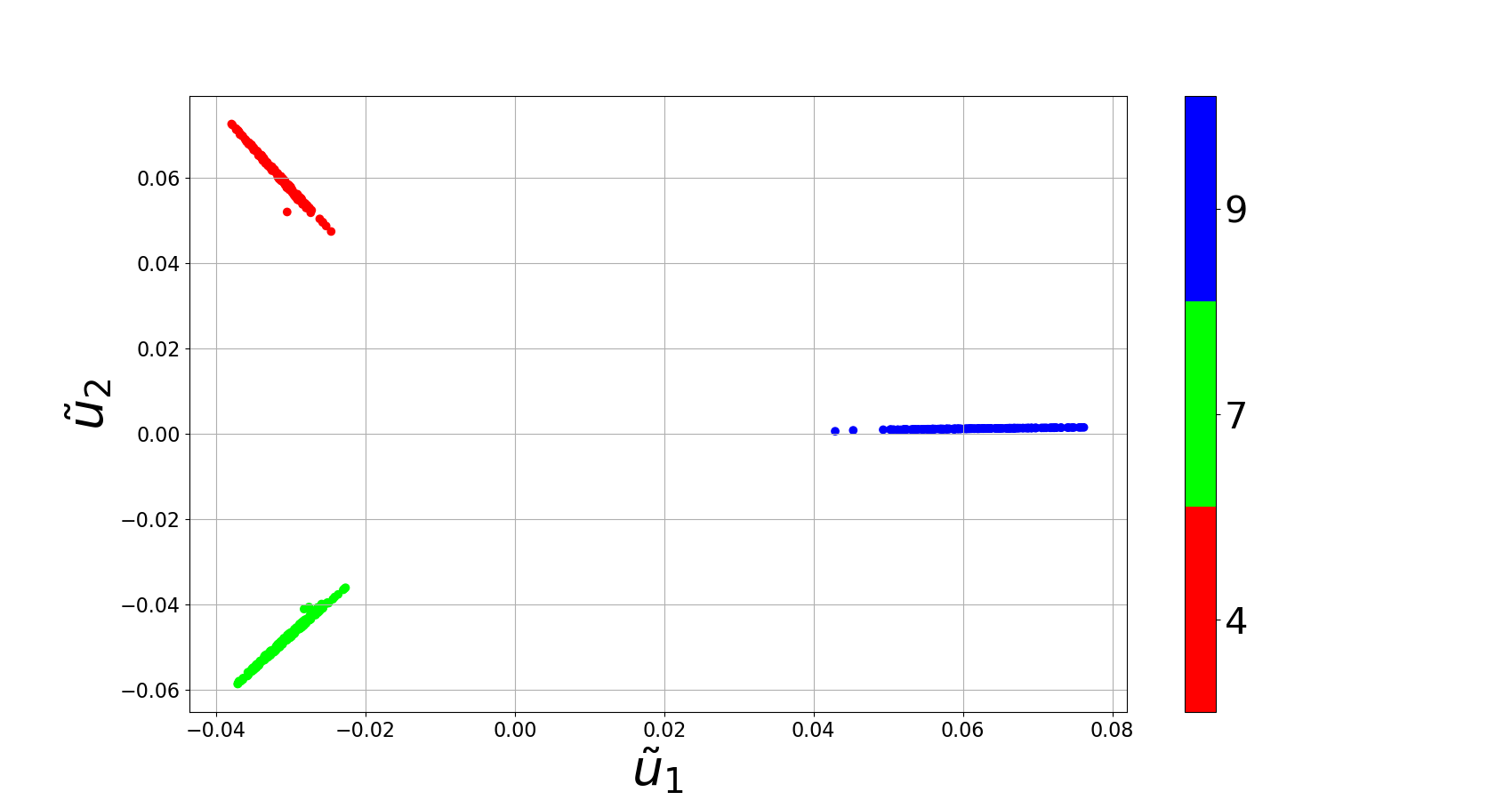}
        \label{fig:tg479_U_sampled_base_before_Tg_iter1500}
    \end{subfigure}
    \hfill
    \begin{subfigure}[t]{0.15\textwidth}
        \centering
        \includegraphics[trim = 40mm 15mm 70mm 20mm, clip=true,width=1\textwidth]{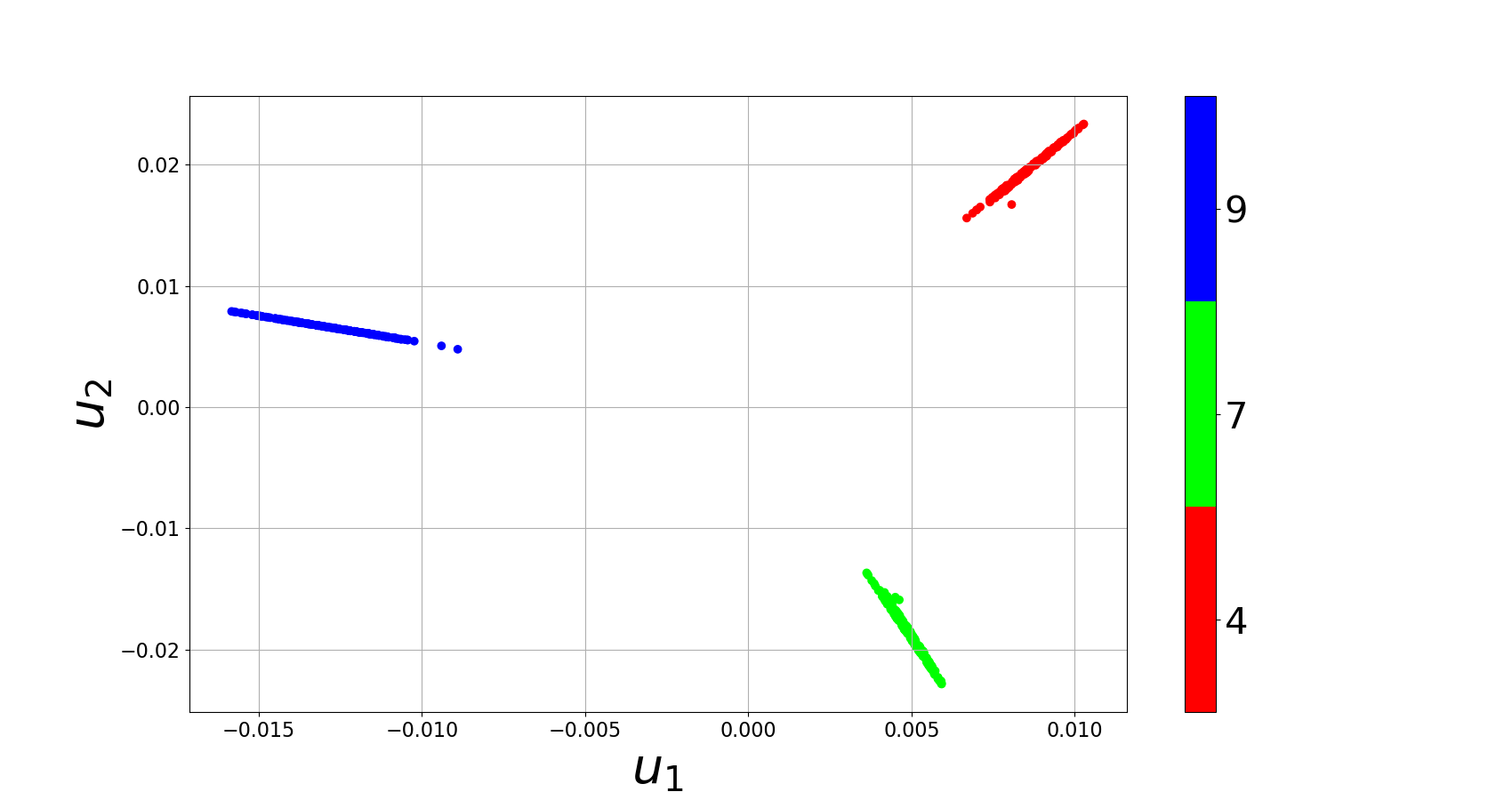}
        \label{fig:tg479_U_sampled_base_after_Tg_iter1500}
    \end{subfigure}
    \hfill
    \begin{subfigure}[t]{0.15\textwidth}
        \centering
        \includegraphics[trim = 40mm 15mm 70mm 20mm, clip=true,width=1\textwidth]{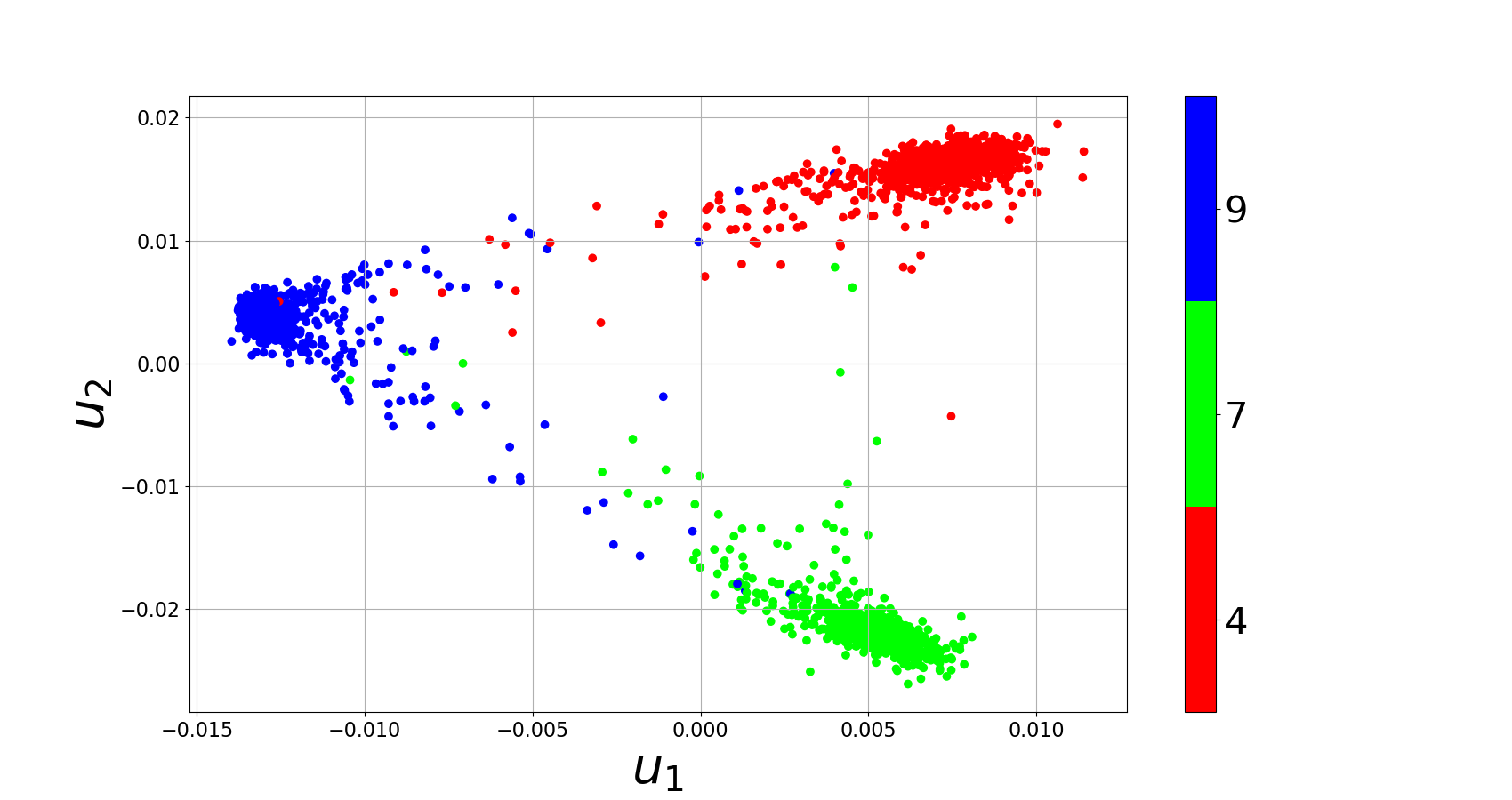}
        \label{fig:tg479_ev_valid_iter1500}
    \end{subfigure}
    \hfill
    \caption{{\bf Features change demonstration} Left column: the analytically calculated spectral embedding of $V^{ref}$ obtained over the features module output. Distortion is a consequence of feature change. Middle column: Spectral embedding of $V^{ref}$ after alignment with $T_G$ . Right column: Network spectral embedding of the test set. Each row represents the embeddings for the 10th, 100th, 500th and 1500th iteration, respectively.
}
    \label{fig:tg_mnist749_embedding_new}
\end{figure}

To illustrate the role of the transformation $T_G$, we show in \cref{fig:tg_mnist749_embedding_new} the results of another experiment using a similar setting. For better visualization, the training is only for three digits of MNIST: 4,7 and 9. The embedding (and visualization) is two dimensional. It can be seen that $T_G$ imposes consistency of the resulting embedding under feature change, allowing convergence and good approximation of the eigenvector space.

\section{Conclusion}
In this paper we introduced BASiS, a new method for learning the eigenspace of a graph, in a supervised manner,  allowing the use of batch training. Our proposed method has shown to be highly robust and accurate in approximating the analytic spectral space, surpassing all other methods with respect to Grassman distance, orthogonality, NMI, ACC and accuracy, over various benchmarks. In addition, we proposed an adaptation of our procedure for learning the eigenspace during iterative changes in the graph metric (as common in neural training). Our method can be viewed as a useful building block for integrating analytical spectral methods in deep learning algorithms. This enables to effectively use  extensive theory and practices available, related to classical spectral embedding.

{\bf Acknowledgements.}
We acknowledge support by the Israel Science Foundation (Grant No.  534/19), by the Ministry of Science and Technology (Grant No. 5074/22) and by the Ollendorff Minerva Center.

{\small
\bibliographystyle{ieee_fullname}
\bibliography{bibli}
}

\end{document}